\documentclass{article}

\usepackage[nonatbib, preprint]{neurips_2024}

\usepackage[utf8]{inputenc} % allow utf-8 input
\usepackage[T1]{fontenc}    % use 8-bit T1 fonts
\usepackage{hyperref}       % h yperlinks
\usepackage{url}            % simple URL typesetting
\usepackage{booktabs}       % professional-quality tables
\usepackage{amsfonts}       % blackboard math symbols
\usepackage{nicefrac}       % compact symbols for 1/2, etc.
\usepackage{microtype}      % microtypography

\usepackage{amsmath}
\usepackage{graphicx}
\usepackage{cleveref}
\usepackage{amssymb}
\usepackage[table]{xcolor}
\usepackage{algorithm}% http://ctan.org/pkg/algorithms
\usepackage{algorithmic}
\usepackage{subfig}

\title{Linear Projections of Teacher Embeddings for Few-Class Distillation}

\author{%
  Noel Loo \\
  MIT \\ 
Computer Science and  \\
  Artificial Intelligence Lab \\
  \texttt{loo@mit.edu} \\
  \And
  Fotis Iliopoulos \\
  Google Research\\
  \texttt{fotisi@google.com} \\
  \AND
  Wei Hu \\
  \\ University of Michigan \\
  Department of Computer Science and \\ Engineering  \\
  \texttt{vvh@umich.edu} \\
  \AND
  Erik Vee \\
  Google Research \\
  \texttt{erikvee@google.com} \\
}

\begin{document}

\maketitle

\begin{abstract}
Knowledge Distillation (KD) has emerged as a promising approach for transferring knowledge from a larger, more complex teacher model to a smaller student model. Traditionally, KD involves training the student to mimic the teacher's output probabilities, while more advanced techniques have explored guiding the student to adopt the teacher's internal representations. Despite its widespread success, the performance of KD in binary classification and few-class problems has been less satisfactory. This is because the information about the teacher model’s generalization patterns scales directly with the number of classes. Moreover, several
sophisticated distillation methods may not be universally applicable or effective
for data types beyond Computer Vision. Consequently, effective distillation techniques remain elusive for a range of key real-world applications, such as sentiment analysis, search query understanding, and advertisement-query relevance assessment. Taking these observations into account,  we introduce a novel method for distilling knowledge from the teacher model's representations, which we term Learning Embedding Linear Projections (LELP). Inspired by recent findings about the structure of final-layer representations, LELP works by identifying informative linear subspaces in the teacher's embedding space, and splitting them into pseudo-subclasses. The student
model is then trained to replicate these pseudo-subclasses.   Our experimental evaluations on large-scale NLP benchmarks like Amazon Reviews and Sentiment140 demonstrate that LELP is consistently competitive with, and typically superior to, existing state-of-the-art distillation algorithms for binary and few-class problems, where most KD methods suffer.
\end{abstract}

\section{Introduction}

While deep neural networks have revolutionized Natural Language Processing~\cite{devlin2018bert, brown2020language}, Computer Vision~\cite{simonyan2014very}, and other fields, their ballooning size and data demands raise challenges. Recent research~\cite{menghani2021efficient} aims to develop efficient models that excel without needing massive datasets or expensive hardware. Knowledge Distillation (KD)~\cite{bucilua,distillation}  is a powerful approach for generating lightweight models by leveraging a large teacher model to guide their training. In its basic form, the student model is trained to replicate the teacher's output probabilities for each training instance. Additionally, several subsequent studies (e.g., \cite{romero2014fitnets, kim2018paraphrasing, passalis2018unsupervised, Ahn2019CVPR, muller2020subclass}) have proposed advanced distillation techniques that go beyond mere output probability matching and focus on encouraging the student to learn the teacher's internal representations.

While KD can significantly improve student model performance in tasks with numerous classes, its impact is less pronounced in binary classification and problems with a smaller number of classes.  As~\cite{muller2020subclass} points out, this is because when we distill knowledge using logits or high-temperature cross-entropy, the information about the teacher model's generalization patterns scales directly with the number of classes. Furthermore, Knowledge Distillation research has primarily focused on Computer Vision, so many sophisticated distillation techniques are not always effective or even suitable for other data modalities like Natural Language. As a result,  effective distillation techniques remain elusive for a range of critical real-world applications, such as sentiment
analysis, search query understanding, and advertisement-query relevance assessment.

\begin{figure}[t]
\vskip 0.0in
\begin{center}
% \includegraphics[width = 0.3\columnwidth]{figures/lelp_clickbait.pdf}
% \caption{Learning with Embedding Projections (LELP) teaches students about subclass structure via linear projections. Learning this structure improves performance over standard knowledge distillation by up to $7.0\%$ on binary CIFAR-100.}
% \label{app:fig:training_curve}

\centering
    \subfloat[]{{\includegraphics[height=3.5cm]{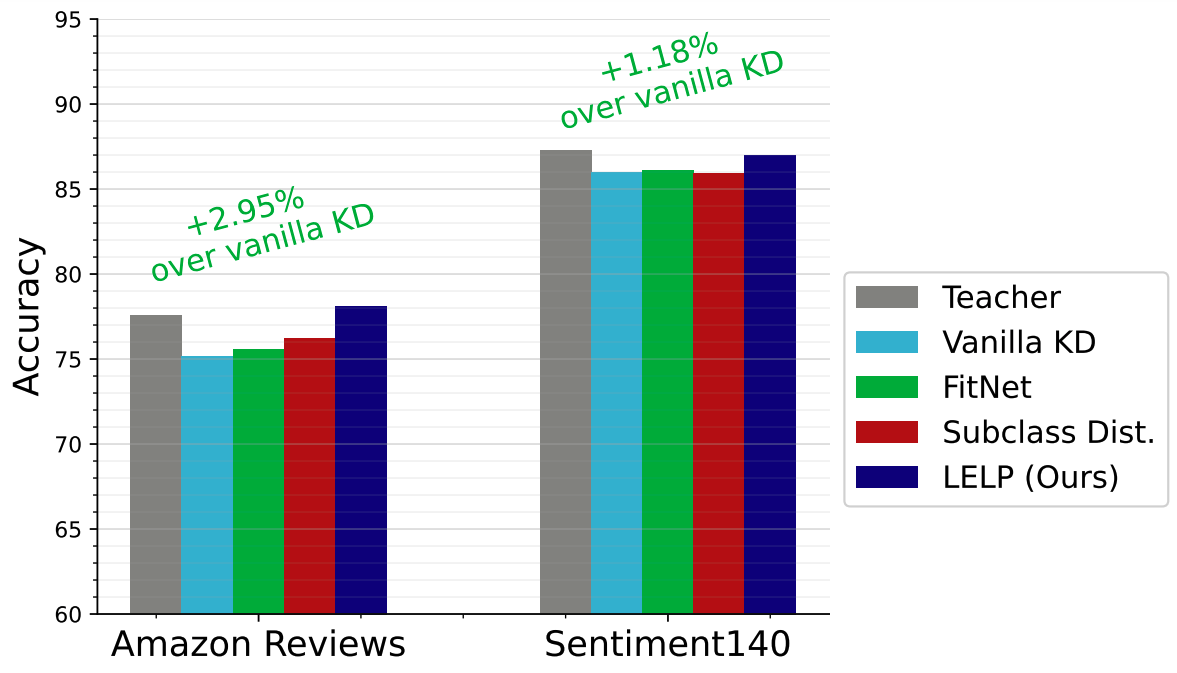} }}%
    \qquad
    \subfloat[]{{\includegraphics[height=3.5cm]{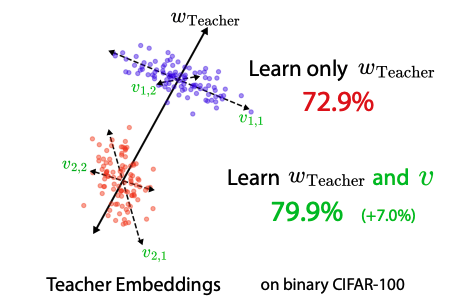} }}%
    \caption{Learning with Embedding Projections (LELP) teaches students about subclass structure (shown by $v_{c,i}$ on (b)) via linear projections. As seen by (a), LELP outperforms existing algorithms over large scale real-world NLP datasets such as Amazon Reviews  (5 classes, 500k examples)  and Sentiment140  (binary, 1.6 million examples) achieving an improvement of 1.85\% and 0.88\%, respectively, over the best baseline. In fact, in the former case, the LELP-trained student outperforms even the teacher, \textbf{which contains over 20x the number of parameters}.
}

    % \label{fig:example}%
\end{center}
\vskip -0.3in
\end{figure}

% Following the seminal works of~\cite{bucilua,distillation}, the technique has gained widespread adoption for efficiently generating high-performing models. This surge in interest is reflected in a multitude of research efforts focusing on both practical applications~\cite{gou2021knowledge, sanh2019distilbert, ruffy2019state} and theoretical analysis~\cite{phuong2019towards, mobahi2020self, menon2020distillation,allen2020towards, borup2021even, harutyunyan2023supervision}.  

% Additionally, several subsequent works (e.g., \cite{romero2014fitnets, kim2018paraphrasing, passalis2018unsupervised, Ahn2019CVPR, muller2020subclass}) have proposed advanced distillation techniques that promote a deeper level of agreement between teacher and student models. These methods move beyond simply matching output probabilities and focus on encouraging the student to learn the teacher's internal representations as well. The intuition here is that these representations contain rich information about the input data~\cite{rumelhart1985learning, levy2014linguistic,olah2017feature} which is not necessarily conveyed to the student model via the teacher's output probabilities.

% especially for classification tasks with a small number of classes~\cite{muller2020subclass}.

In light of these observations, in this paper we propose a novel approach for capturing information from the teacher's representations, which we call Learning Embedding Linear Projections (LELP). At a high level, it works by extracting knowledge form the teacher's last-layer representations (embeddings) and converting it into pseudo-subclasses via linear projections. The student model is then trained on these pseudo-subclasses, using a single unified cross-entropy loss. 

Our approach leverages recent findings about the structure of final-layer representations (embeddings) in deep learning models and, in particular, aspects of the phenomenon known as \emph{Neural Collapse}~\cite{papyan2020prevalence, fang2021exploring, yang2023neurons}. Similar in spirit to Subclass Distillation~\cite{muller2020subclass}, which uses a modified and retrained teacher model to identify hidden patterns, our method achieves improved student performance, particularly in finetuning Language Models for  tasks with a few number of classes. Crucially though, and unlike Subclass Distillation, our approach does not require any retraining of the teacher model.  Finally,  a key advantage of LELP is its flexibility in bridging diverse model architectures, making it uniquely versatile for teacher-student learning scenarios. 

Our contributions can be summarized as follows:

\begin{enumerate}

    \item Motivated by recent insights into the Neural Collapse phenomenon, we demonstrate that the invention of pseudo-subclasses through unsupervised clustering of teacher embeddings can enhance distillation performance in binary and few-class classification tasks. However, the efficacy of this approach is contingent upon the specific clustering algorithm employed.

    \item Through empirical evaluation, we observe that carefully calibrated linear projections consistently achieve high performance, leading us to introduce LELP: a novel method for enhancing Knowledge Distillation. LELP is modality-independent, producing particularly strong results in NLP tasks and situations where the teacher and student  architectures differ.
    
    \item Empirical evaluations on large-scale  NLP benchmarks like Amazon Reviews  (5 classes, 500k examples, achieving an improvement of 1.85\% over the  best baseline)  and Sentiment140  (binary, 1.6 million examples, showing a 0.88\% improvement over the  best baseline) validate that LELP is consistently competitive with, and typically superior to, existing SOTA distillation algorithms for binary and few class problems, where most KD methods suffer.

    \item We show that LELP possesses several advantageous characteristics, including conveying a large amount of information per example (data efficiency), converging faster than Vanilla KD, and can provide substantial improvements in semi-supervised KD scenarios.
\end{enumerate}

\subsection{Organization of the paper}

In Section~\ref{sec:related_work} we discuss related work, including the baselines we considered in this work and the Neural Collapse phenomenon.  In Section~\ref{sec:our_method} we present our method.  In Section~\ref{sec:experimental_evaluation} we present our experiments. Specifically, in Section~\ref{unsupervised_clustering_comparison} we present experiments demonstrating that inventing pseudo-subclasses via clustering of the teacher's model's embeddings can improve distillation effectiveness, and in Section~\ref{comparison_with_baselines} we compare LELP with other distillation baselines. In Appendix~\ref{app:broader_impact}   we discuss the broader impact of our work.   In Appendix~\ref{app:binary_subclass_structure_results} we present our experimental results corresponding to binary classification tasks with subclass structure in detail. In Appendix~\ref{app:ablations} we provide ablations on the design choices behind LELP. In Appendix~\ref{app:NLP_datasets_details} we give a detailed description of the NLP datasets we used.   In Appendix~\ref{app:multiclass} we present results regarding the applicability of LELP to multiclass classification tasks with a moderate number of classes.
In Appendix~\ref{app:properties} we study additional properties of LELP, including its relative performance in the \emph{semi-supervised setting}~\cite{chen2020big,stanton2021does,IKBMTV22, RAD,kontonis2023slam}, its data efficiency and its training speed.  In Appendix~\ref{app:alg_details} we provide pseudo-code for our method.  Finally, in Appendix~\ref{sec:implementation_details} we provide implementation details.

\section{Related Work}
\label{sec:related_work}

\textbf{Knowledge Distillation.}
Most of the literature on Knowledge Distillation has been focused on the \emph{fully supervised} setting, i.e., when distillation is performed on the labeled training data of the teacher model rather than on new, unlabeled data — see e.g. the original paper~\cite{distillation}. Specifically, when training the student one typically uses a convex combination of the standard cross-entropy loss $\mathcal{L}_{\mathrm{CE}}$ with respect to the ground-truth labels, and the 
\emph{distillation loss} $\mathcal{L}_{\mathrm{distill}}$:
\begin{align}\label{student_loss}
\mathcal{L}_{\mathrm{student}} =  \alpha \mathcal{L}_{\mathrm{CE}} + (1- \alpha) \mathcal{L}_{\mathrm{distill}}.
\end{align}
In the original paper~\cite{distillation} the distillation loss encourages the student model to mimic to replicate the teacher's output probabilities, potentially after scaling the logits of both models using a temperature-scalar. (The value of the temperature is typically larger than $1$, and it is used to emphasize the differences between the probabilities of wrong answers that would all be very close to zero at temperature $1$.)

In addition to focusing on the teacher model's final outputs, follow-up methods, like those proposed in~\cite{romero2014fitnets, Ahn2019CVPR}, also consider the teacher model's internal representations, often in the form of its embeddings. These methods encourage the student model to mimic not only the teacher's final predictions but also its internal representations. The simplest, yet often highly effective, example of this approach is \emph{Embedding Distillation}~\cite{romero2014fitnets}. In this method, an additional term, called the ``embedding-loss term'', is added to the distillation loss. This term measures the difference between the embeddings produced by the teacher and student models. The weight of this term can be adjusted using a hyperparameter. Specifically, one adds the term
\begin{align}
\mathcal{L}_{\mathrm{Embedd}} = \frac{1}{n} \sum_{i=1}^{n} \|f^T(x_i) - W f^S(x_i)  \|_2,
\end{align}
where $x_1, \ldots, x_n$ denotes the current batch of examples, $f^T(x_i), f^S(x_i)$ the embeddings of the teacher and student model  corresponding to example $x_i$,  respectively, and $W$ is a learnable projection matrix to match the dimension of the teacher/student learned during the distillation phase. \cite{romero2014fitnets} proposes ways of modeling and pretraining $W$ (and also potentially distilling from other teacher-layers), while~\cite{Ahn2019CVPR} proposes losses that minimize the mutual information between the teacher and student embeddings (instead of considering the $\ell_2$ loss). Finally, Relational Knowledge Distillation~\cite{park2019relational} transfers mutual relations of data examples instead, e.g., they authors introduce a loss that penalizes structural differences in ``anglewise'' relations. Notably, Relational KD can naturally handle mismatches in teacher-student embedding dimensionality without introducing learnable projections.

The Subclass Distillation method~\cite{muller2020subclass} presents an approach that is closely related to ours. Here, the teacher is forced to divide each class into many pseudo-subclasses that it invents via appropriate training, and then the student is trained to match the subclass probabilities. The method is designed for few-class classification, excelling in such scenarios. Our method shares Subclass Distillation's use of invented subclasses for student knowledge transfer, but eliminates the need for teacher retraining and extensive hyperparameter tuning. (Note that the process of optimizing hyperparameters for Subclass Distillation necessitates retraining the teacher model repeatedly with varying loss function configurations, as outlined in equations (7) and (8) of the reference paper by Muller et al. (2020). However, this iterative retraining becomes \emph{excessively computationally intensive and impractical} when dealing with large teacher models.) As we show in Section~\ref{sec:experimental_evaluation}, LELP achieves performance that is always on par with, and typically exceeding, Subclass Distillation.

In our study we  show that LELP is consistently competitive with the above methods, and that it is typically superior to them when it comes to  classification problems with a few number of classes (especially in the NLP domain). We also explore two more baseline methods that have found success in Vision tasks: Contrastive Representation Distillation (CRD)~\cite{tian2019contrastive} and Decoupled Knowledge Distillation (DKD)~\cite{zhao2022decoupled}. While CRD excels in Vision multi-class tasks, its reliance on data augmentation for contrastive learning limits its applicability to other domains. Indeed, in Vision classification tasks the standard ``contrastive-learning'' approach would be to generate variations of every image (e.g, via rotations), and then consider as positive any pair of examples (original or modified) that come from the same source. However, data augmentation of, say Natural Language examples, is not straightforward and this impairs the performance of the method. (See Appendix~\ref{sec:implementation_details} for more details.)  DKD has also demonstrated efficacy in multi-class Vision tasks, but its performance is notably compromised in scenarios with fewer classes, rendering it inappropriate for the specific focus of our investigation. In particular, in the context of binary classification, DKD is mathematically equivalent to the standard Vanilla KD approach. We demonstrate that LELP significantly outperforms both CRD and DKD in text classification tasks.

% Finally, in Appendix~\ref{app:properties} we study additional properties of LELP, including its relative performance in the \emph{semi-supervised setting}~\cite{chen2020big,stanton2021does,IKBMTV22, RAD,kontonis2023slam}, i.e., the setting where distillation is performed over a large pool of new, unlabeled examples, and which is the setting which often sees the greatest benefit of distillation.

\textbf{Neural Collapse.}
\cite{papyan2020prevalence} identified a phenomenon called \emph{Neural Collapse} that occurs in the final training phase (after achieving zero training error) of deep neural networks. This phenomenon refers to specific structural properties observed in the last layer's representations. The simplest of these properties is known as \emph{variability collapse}. In a nutshell,  as a  network is trained extensively, it tends to group together all training samples with the same label (almost) into a single point in its final layer. 

Our approach is inspired by a recent paper~\cite{yang2023neurons} which provides evidence that further refines the above description. In particular, it shows that while the final layer representations appear collapsed, they retain crucial fine-grained structure. This structure, though subtle, accurately reflects the inherent characteristics of the input distribution. As an illustrative example, the authors consider a model trained to classify images in the CIFAR-10~\cite{krizhevsky2009learning} dataset using only 5 coarse-graned labels (by combining two classes into one superclass). Remarkably, even after this training, they can still recover the original, more detailed 10-class system by applying unsupervised clustering techniques to the model's internal representations.   In light of this intriguing observation, and  motivated by the findings of~\cite{muller2020subclass} that creating meaningful pseudo-subclasses aids student knowledge transfer, we present an unsupervised technique to extract knowledge from teacher embeddings using linear projections to form pseudo-subclasses. Our approach, which we compare to other clustering methods (including the one in~\cite{yang2023neurons}) in Table~\ref{tab:binary-subclass-structure} in Section~\ref{unsupervised_clustering_comparison}, consistently shows superior performance.

\begin{figure*}[t]
\vskip 0.0in
\begin{center}
\includegraphics[width = 1.00\linewidth]{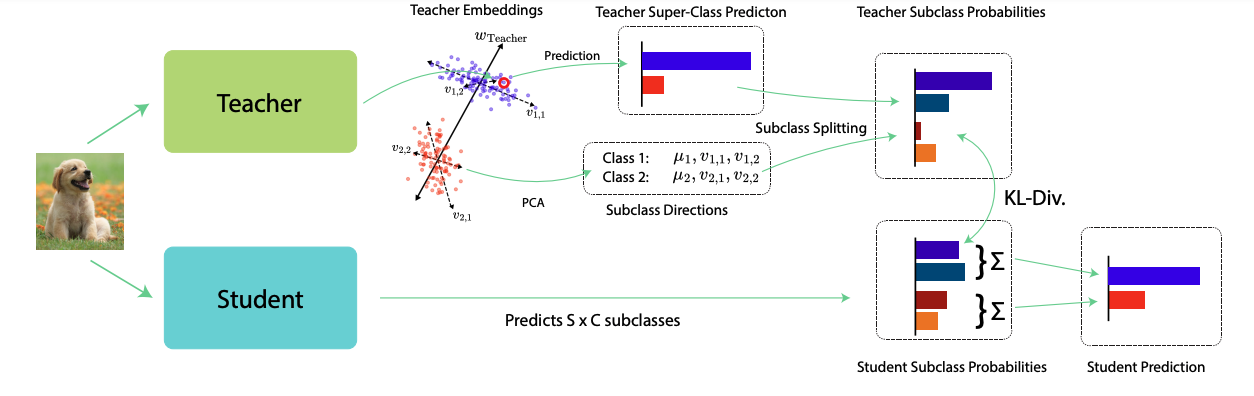}
\caption{Schematic of the Learning with Embedding Projections (LELP) algorithm. LELP decomposes Teacher predictions into subclasses via a PCA decomposition, and trains a student on these subclasses. For predictions, subclasses are summed together back into their original classes.}
\label{app:fig:training_curve}
\end{center}
\vskip -0.25in
\end{figure*}

\section{Learning from Embedding Linear Projections (LELP)}
\label{sec:our_method}

Since large models develop embeddings that hold information not captured in their output probabilities \cite{yang2023neurons}, we aim to transfer this knowledge to student networks while adhering to three  desiderata: 

\textbf{Modality-independent}. We aim to develop a data-agnostic method, meaning its performance remains consistent regardless of the underlying data type. The method should not be reliant on characteristics specific to certain data modalities (for instance, the ability to easily perform data augmentation).

\textbf{Compatibility between differing Student/Teacher architectures}. Embedding Distillation, the most straightforward method for learning the teacher's embedding information, only works out-of-the-box when the student and teacher have matching embedding dimensions. Otherwise, a learnable projection layer is required to match the student/teacher embedding dimensions, which can often harm performance (and even increase latency in certain cases). Therefore, we aim for a method that is embedding dimension agnostic.

\textbf{No Retraining the Teacher Model}. When the teacher model is significantly larger than the student model, retraining it can be prohibitively expensive. Methods which require a modified teacher training pipeline are significantly more costly to tune hyperparameters for, as not only do the teacher models have to be retrained multiple times, but in addition the student hyperparameters need to be checked against varying teacher hyparameters. This precludes methods such as Subclass Distillation \cite{muller2020subclass}, which contains several teacher training hyperparameters, and can even hurt the teacher's performance.

% \textbf{Minimal Modification of the Training pipeline} Our method should fit into the common single-objective optimization framework with minimal modification. Other methods of extracting embedding information such as FitNet \citep{romero2014fitnets} or VID \citep{Ahn2019CVPR} require more careful scheduling of learning objectives, or more complex minimax objectives which are harder to implement in many practical training pipelines.

% This name is kinda bad but i cna't think of anything better
With these requirements in mind, we present Learning Embedding Linear Projections (LELP), which extracts the information in the teacher's embedding layer into pseudo-subclasses, on which the student is trained in a single unified cross-entropy loss. To motivate our method, we refer to the toy model of the learned teacher embeddings shown in Figure~\ref{app:fig:training_curve}. Here, data points cluster around their respective class averages, but each cluster exhibits internal structure that holds semantically meaningful information. This structure differentiates individual items within the same class and has proven valuable for identifying subclasses when they exist~\cite{muller2020subclass}. Additionally, \cite{yang2023neurons} shows that these subclasses can be separated using a linear probe (following an appropriate clustering of the embeddings space), which motivates the use of linear projections as a means of extracting further information for distillation.

LELP extracts this embedding information into a single classification loss in three steps. Firstly, we identify meaningful linear subspaces in the teacher embedding space (\cref{sec:methods_pca}). Secondly, we  project the teacher embeddings into these subspaces, and use them to form pseudo-subclasses, expanding the number of classes from $C$ to $S\times C$, where $S$ is the number of linear projections we employ per class (\cref{sec:methods_pseudo_class_splitting}). Finally, we perform standard knowledge distillation, where additionally the student network's final layer outputs probabilities for $S\times C$ classes (\cref{sec:methods_kd_with_subclasses}).

\subsection{Identifying Informative Linear Subspaces}
\label{sec:methods_pca}
The first step in LELP requires identifying linear subspaces in the teacher embedding space which contain useful information for each class cluster. Specifically, let $\{x^c_i\}_{i=1}^{N_c}$, be the $N_c$ training points in the dataset belonging to class $c$, and let $\{h^c_i\}_{i=1}^{N_c}$ be the corresponding teacher embeddings in $\mathbb{R}^D$, that is, $h^c_i = h^\text{Teacher}(x^c_i)$, where $h^\text{Teacher}(x)$ is the teacher feature extractor. We want to find the $S$ most informative linear directions in $\{h^c_i\}_{i=1}^{N_c}$. In absence of further knowledge of the structure of $\{h^c_i\}_{i=1}^{N_c}$, we opt to take the $S$ top PCA directions. This PCA decomposition yields for each class a class mean $\mu_c$ and $S$ top PCA directions: $\{v_{c, 1},v_{c, 2}, \dots, v_{c, S} \}$. Two remarks are in order.

First, observe that such a PCA can contain ``redundant'' information which is already captured in the teacher's output weights. Specifically, if the teacher's output weights are $\{w_i, \dots, w_c\}$, standard knowledge-distillation will already contain all the information along these directions, meaning that further information in these directions from $v_{c, i}$ is unnecessary. Therefore, we have found that it often helps to first project $\{h^c_i\}_{i=1}^{N_c}$ onto the null-space of the teacher weights $\{w_i, \dots, w_c\}$ before performing the PCA. Of course, in principle, the null-space could be trivial, in which case we do not apply this step. Typically though, we are working in the regime where $D \ge S + C$, so this step does apply. In particular, we have not  run into the problem in the experiments of this study.

A second consideration is the imbalance of variance between PCA directions:  projections onto the $v_{c,1}$ directions to contain the most variances and each subsequent will contain less. If there is significant imbalance, we find that this can lead to poor performance so we have found it is useful to apply a random rotation on the PCA directions so the variance in each direction is equal. Specifically, we use $\tilde{V}_{c} = Q V_{c}$, where $V_{c} = [v_{c, 1}, \dots, v_{c, S}]\in \mathbf{R}^{S\times D}$ is our PCA vectors concatenated and $Q \in \mathbb{R}^{S \times S}$ is a random orthonormal matrix. $\tilde{V}_{C} = [\tilde{v}_{c, 1}, \dots, \tilde{v}_{c, S}]$ contains our random rotated PCA directions, which span the same space as $V$, but each direction has the same variance in expectation. 

 In Appendix~\ref{app:ablations} we perform ablations where we compare applying LELP  to simply applying PCA or Random Projections, in order to show that the above observations can indeed be beneficial in terms of the student's performance. Algorithm pseudo-code for the step of identifying the linear subspaces is provided in Algorithm~\ref{alg:embedding_pca} in \cref{app:alg_details}.

The cost of performing the PCA is $O(N_c D^2 + D^3)$ in time and $O(D^2)$ in memory, where $D$ is the embedding dimension. In practice, the $O(N_c D^2)$ associated with forward-passing the dataset through the teacher network is the most costly compared to the $O(D^3)$ PCA cost for practical embedding dimensions, making the practical cost of the computation $O(N)$ where $N$ is the training dataset size.

\subsection{Splitting into pseudo-subclasses}
\label{sec:methods_pseudo_class_splitting}
 Given the set of $C$ class embedding means and $S$ PCA vectors per class, we now describe how we split the $C$ classes into $SC$ subclasses. Let $p_c^\text{Teacher}$ be the teacher class probability of class $c$, with temperature parameter $\tau$. That is,
\begin{align*}
    p_c^\text{Teacher} = \frac{e^{z_{c}/\tau}}{\sum_{i = 1}^C e^{z_{i}/\tau} },   
\end{align*}
where $z_c$ are teacher logits, i.e. $z_c = w^{\text{Teacher}\intercal}_c h - b_c$, with $w^{\text{Teacher}}_c$ the teacher weight for class $c$ and $b_c$ the bias. We split this into $S$ subclass probabilities ${p_{c, 1}, \dots p_{c, S}}$ where 
\begin{align*}
    p_{c, s}^\text{Teacher} = p_c^\text{Teacher} * \frac{e^{z_{c, s}/\beta}}{\sum_{j = 1}^S e^{z_{c, j}/\beta} }.   
\end{align*}
where $z_{c, s} = \tilde{v}_{c, s}^\intercal (h - \mu_c)$. That is, we perform a tempered softmax over \textit{subclass logits} $z_{c, s}$, where $z_{c, s}$ are given by the PCA decomposition coordinates for that class. $\beta$ is our subclass tempering parameter which is a hyperparameter in our method. We refer to this subclass splitting algorithm as $\texttt{subsplit}$, which takes as input the teacher embedding $h$, the PCA direction and mean vectors $\tilde{V}$ and $M$, the teacher final layer weights $W$, and temperature $\beta$. Pseudocode is in Appendix~\ref{app:alg_details}.

\subsection{Knowledge Distillation with subclasses}
\label{sec:methods_kd_with_subclasses}

Finally, we perform standard knowledge distillation with our new $SC$ probabilities $p_{c, s}$. This requires a straightforward modification to the student architecture, in which it outputs $SC$ classes as opposed to the standard $C$ classes. We applying the standard tempered Knowledge Distillation loss as prior work, using the same temperature $\tau$ used to generate $p_{c,s}$:
\begin{align*}
    \mathcal{L}_{\mathrm{LELP}} = \tau ^2 \mathcal{D}(p_{c, s}^{\text{Teacher}}||p_{c, s}^{\text{Student}}),
\end{align*}
where $\mathcal{D}$ is a standard classification loss like Cross-Entropy or KL-Divergence. (In all our experiments in this paper we  use the latter.)

With $p_{c, i}^{\text{Student}}$ using the same temperature parameter $\tau$:
\begin{align*}
    p_{c,s}^{\text{Student}} = \frac{e^{z_{c, s}^{\text{Student}}/\tau}}{\sum_{i = 1}^C \sum_{j = 1}^S e^{z_{i, j}^{\text{Student}}/\tau} }.   
\end{align*}

At test time, we simply take class probabilities to be the sum over subclass probabilities: $p_{c}^{\text{Student}} = \sum_{j = 1}^S p_{c,j}^{\text{Student}}$ and take the prediction to be the class with largest probability. By putting embedding information directly into a unified classification loss, our algorithm avoids careful balancing of training objectives such as with Embedding Distillation, in which one must carefully tune the embedding loss coefficient. Furthermore, our method is minimally invasive, requiring only minor changes to the student network and the training pipeline relative to other effective Knowledge Distillation techniques. In particular, our method avoids the need for pretraining steps of learnable projections, unlike FitNet and VID, and it also doesn't require a memory buffer (sometimes used for CRD), or retraining the teacher model like Subclass Distillation. All put together, our algorithm is given in Algorithm~\ref{alg:LELP_train}.

\section{Experimental Evaluation}
\label{sec:experimental_evaluation}

In this section we present our experimental results. In Section~\ref{the_setup} we describe our experimental setup.  Section~\ref{unsupervised_clustering_comparison} presents experiments showcasing that generating pseudo-subclasses via unsupervised clustering of the teacher model's embeddings can improve distillation effectiveness. In Section~\ref{comparison_with_baselines} we compare LELP with other distillation baselines.

\subsection{The Setup}
\label{the_setup}
We focus our experiments on a variety of  classification tasks, using a standard distillation setup as in \cite{distillation}. In order to focus solely on the effect of the distillation loss of each method, we always set $\alpha = 0$ in ~\eqref{student_loss}. This reduces the variance between methods which may have different optimal values of $\alpha$, and reduces the hyperparameter search space. Furthermore, in the important case of the semi-supervised setting one does not have access to ground-truth labels.

\textbf{The Architectures  } Given that the specific combination of student and teacher architectures is known to influence the effectiveness of knowledge distillation, we have chosen to evaluate various student-teacher pairings to ensure the robustness of our method.

For Vision datasets, the ``ideal" distillation scenario is given by distilling ResNet-92 to ResNet-56, where both architectures and embeddings dimensions match ($D = 256$). We also consider the case of a smaller and a larger dimension with ResNet-92 ($D=256$) and MobileNet with width and depth multiplier equal to 2 ($D=2048$), respectively, distilling to MobileNet $(D=1024)$. The latter cases address scenarios where the embedding dimensions differ, with one scenario involving the same architecture family and the other involving different architectures.  For smaller NLP datasets (LMRD, GLUE/cola, and GLUE/sst2), we consider distillation from an ALBERT-Large model ($D=1024$) to an ALBERT-Base model ($D=768$). For the larger scale NLP datasets (based on Amazon US reviews and Sentiment 140) we consider distillation from an ALBERT-XXL model $(D=4096)$ and an ALBERT-XL model $(D=2048)$ to (i) an ALBERT-Base model ($D=768$); (ii) and a two-layer-MLP of width $(D=4096)$ that operates over representations generated by a (frozen) sentence-T5 encoder model of 11 billion parameters \cite{T5_sentence}. The latter case addresses the scenario involving different teacher-student architectures  but with the same embedding dimension. (Using a pre-trained, frozen large-scale encoder model to generate representations is a common-in-practice approach when one needs multiple lightweight models for different classification tasks.)

\textbf{The Baselines  }
To establish a baseline performance, we chose well-known distillation approaches. In particular, we consider Standard Training of the model with the ground-truth labels and the cross-entropy loss, Vanilla Distillation with temperature-scaling as described in the original paper of~\cite{distillation}, the  Embedding Distillation method as described in Section~\ref{sec:related_work}, the general FitNet~\cite{romero2014fitnets} approach\footnote{Embedding Distillation as described in Section~\ref{sec:related_work} can be thought of as an instance of the FitNet framework.} , the Variational Information Distillation for Knowledge Transfer (VID) framework~\cite{Ahn2019CVPR}, the Relational KD approach~\cite{park2019relational}, Contrastive Representation Distillation CRD~\cite{tian2019contrastive}, Decoupled Knowledge Distillation~\cite{zhao2022decoupled} and Subclass Distillation~\cite{muller2020subclass}. As previously noted, DKD is functionally identical to Vanilla KD in the context of binary classification tasks. Consequently, for such tasks, \emph{reported values for DKD and Vanilla KD are congruent}. For all baselines we perform a grid search over their relevant parameters and report the performance (test-accuracy and  standard deviation over three trials) of the best configuration, with hyperparameters given in Appendix~\ref{hyperparameters_chosen}. The best forming algorithm is shown in bold, and the second best underlined. Additionally, for every scenario in our tables, we display the average improvement of LELP compared to: (i) the top-performing baseline; (ii) the best baseline excluding  Subclass Distillation; (iii) Vanilla Distillation. We report the second comparison for a couple of reasons. Firstly,  optimizing Subclass Distillation's hyperparameters is significantly more costly compared to the other baselines, and it can be challenging in real-world situations (due to the requirement of retraining and storing multiple teacher models). Second, the accuracy of the teacher model in Subclass Distillation usually differs from the one used for LELP (and the other baselines). Therefore, comparing them directly might not be entirely fair.

\begin{figure}[t]
\vskip 0.0in
\begin{center}
% \includegraphics[width = 0.3\columnwidth]{figures/lelp_clickbait.pdf}
% \caption{Learning with Embedding Projections (LELP) teaches students about subclass structure via linear projections. Learning this structure improves performance over standard knowledge distillation by up to $7.0\%$ on binary CIFAR-100.}
% \label{app:fig:training_curve}

\centering
    % \subfloat[]{{\includegraphics[height=4cm]{figures/chart_res_res.pdf} }}%
    \subfloat[]{{\includegraphics[height=3.5cm]{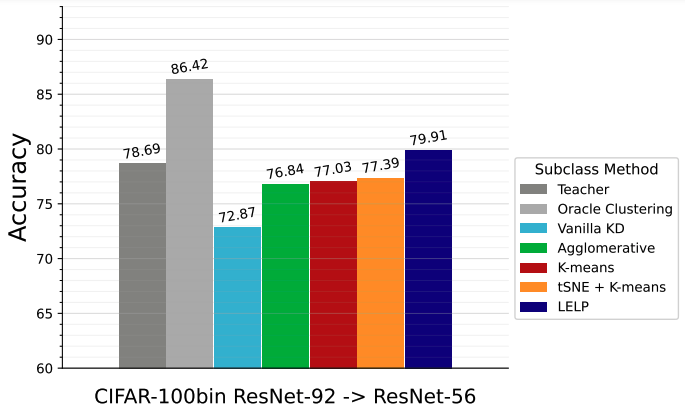} }}%
    \qquad
    \subfloat[]{{\includegraphics[height=3.5cm]{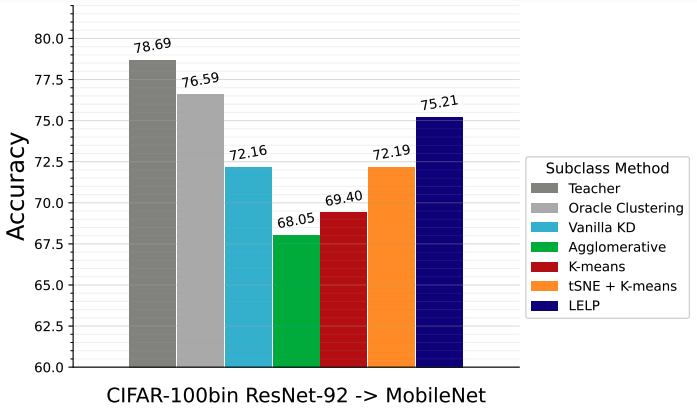} }}%
    \caption{The effectiveness of different clustering techniques for creating pseudo-subclasses during knowledge distillation from a ResNet-92 teacher to (a) a ResNet-56 and (b) a MobileNet student on the binarized CIFAR-100 dataset is presented.  ``Oracle Clustering'', where subclass structure is known a priori, serves as an upper bound and notably surpasses all other methods, even exceeding the teacher's performance in the ResNet-56 case. Among practical methods (i.e., those discovering subclass structure), LELP exhibits superior performance.  Agglomerative and K-means clustering do not consistently outperform vanilla knowledge distillation, demonstrating the dependence of pseudo-subclass effectiveness on the chosen clustering algorithm. }
\label{fig:clustering_methods}  
\end{center}
\vskip -0.3in
\end{figure}

% \begin{figure*}[t]
% \vskip 0.0in
% \begin{center}
% \includegraphics[scale=0.4]{figures/chart_res_res.pdf}
% \caption{The performance of various clustering methods for inventing pseudo-subclasses when distilling a ResNet-92 to a ResNet-56 in CIFAR-100 bin. We observe the following: (i) ``Oracle Clustering'' which represents the idealized scenario where subclass structure is known in advance, outperforms all other methods, and even the teacher model, underscoring the potential power of inventing pseudo-subclasses. (ii) LELP is the best performing approach among the ``practical ones'', i.e., the ones which they do not know the subclass structure in advance and try to discover it. (iii) In this scenario all methods significantly outperform Vanilla KD, but th }
% \label{app:fig:clustering_chart}
% \end{center}
% \vskip -0.25in
% \end{figure*}

\begin{figure}[htbp]
    \centering
    \begin{tabular}{cccc} % 4 columns
        % \begin{minipage}{0.23\textwidth}
        %     \centering
        %     \includegraphics[width=\textwidth]{figures/cifar10bin_teacher_embeddings.png}
        % \end{minipage} &
        \begin{minipage}{0.23\textwidth}
            \centering
            \includegraphics[width=\textwidth]{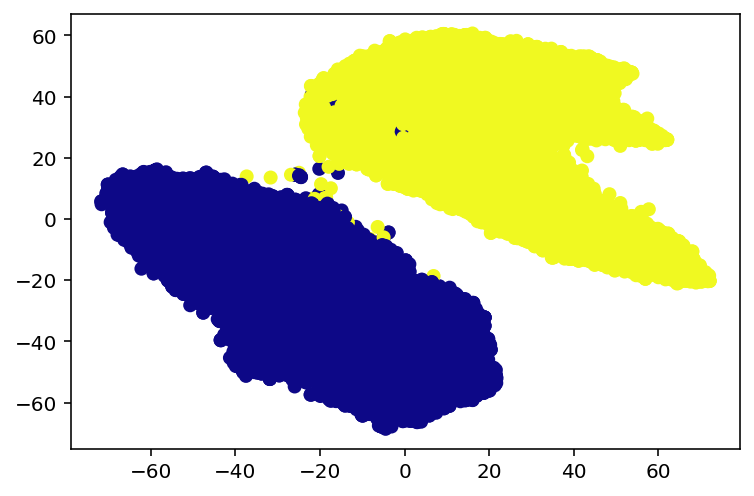}
        \end{minipage} &
        \begin{minipage}{0.23\textwidth}
            \centering
            \includegraphics[width=\textwidth]{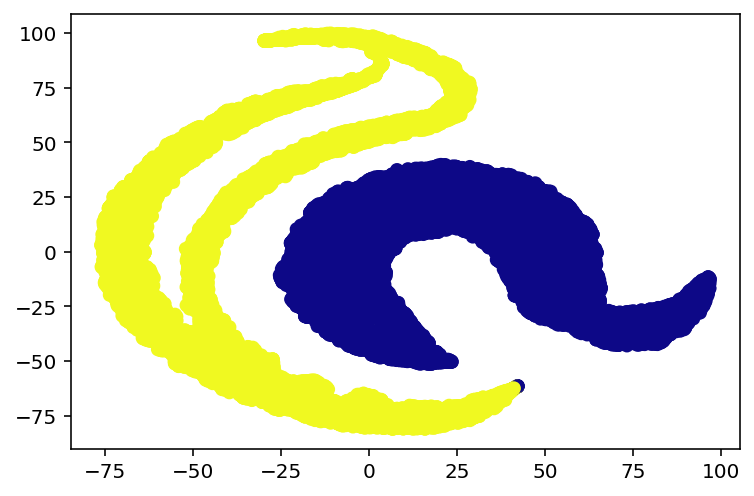}
        \end{minipage} &
        \begin{minipage}{0.23\textwidth}
            \centering
            \includegraphics[width=\textwidth]{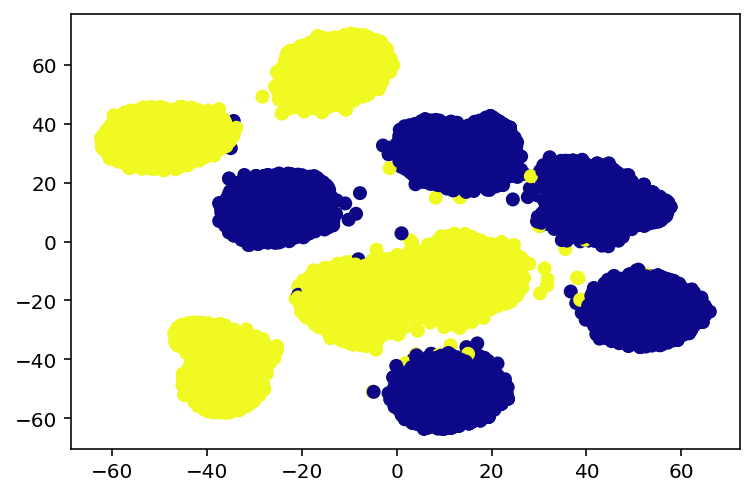}
        \end{minipage} \\
        
        % \begin{minipage}{0.23\textwidth}
        %     \centering
        %     \includegraphics[width=\textwidth]{figures/cifar10bin_teacher_sub.png}
        %     \caption*{Teacher}
        % \end{minipage} &
        \begin{minipage}{0.23\textwidth}
            \centering
            \includegraphics[width=\textwidth]{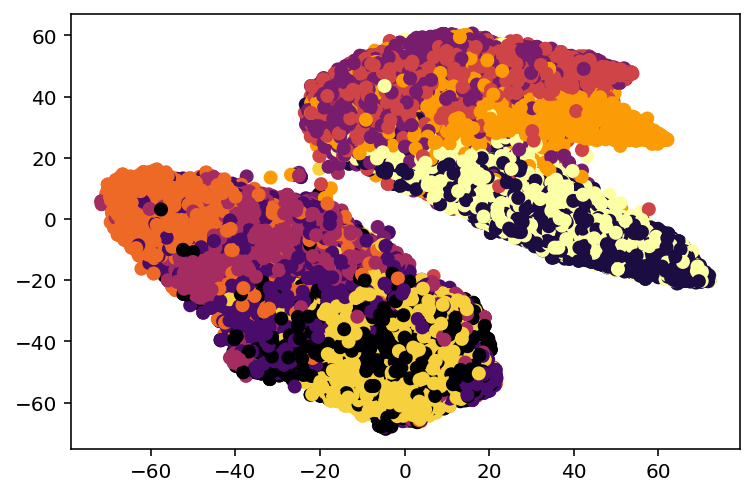}
            \caption*{Vanilla KD}
        \end{minipage} &
        \begin{minipage}{0.23\textwidth}
            \centering
            \includegraphics[width=\textwidth]{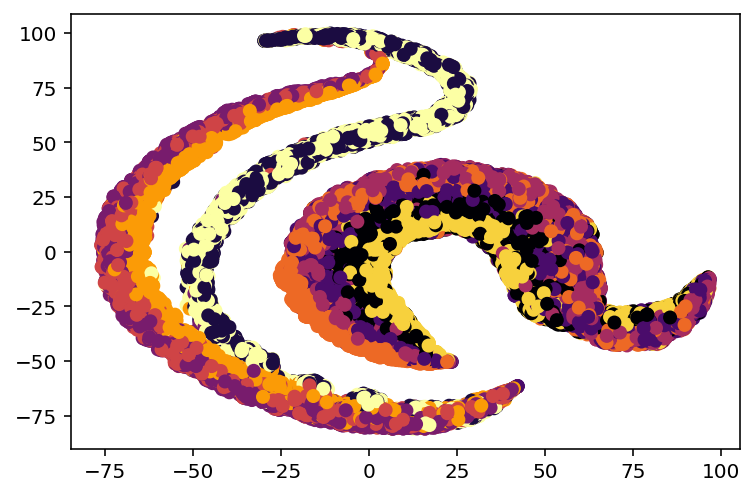}
            \caption*{LELP}
        \end{minipage} &
        \begin{minipage}{0.23\textwidth}
            \centering
            \includegraphics[width=\textwidth]{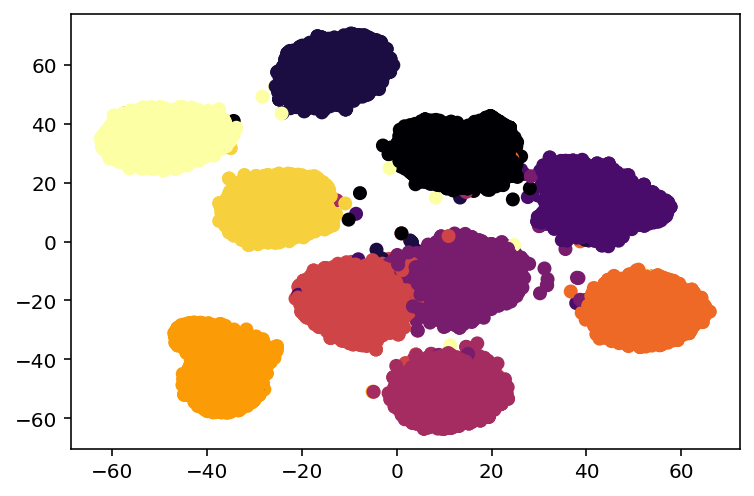}
            \caption*{Oracle Clustering}
        \end{minipage} \\
    \end{tabular}
    \caption{The t-SNE visualization of the feature embeddings learned by student models during knowledge distillation from a ResNet-92 teacher to a ResNet-56 student on the binarized CIFAR-10 dataset. The first row depicts embeddings colored according to the two primary classes of the task, while the second row uses color to represent the underlying subclasses.  Notably, the student model trained with the proposed LELP method exhibits a more intricate embedding structure compared to the student trained with standard Vanilla KD. Furthermore, the embeddings of the student trained via the ``ideal'' Oracle Clustering approach clearly delineate the underlying subclass structure, providing insight into its ability to surpass even the teacher's performance.}
    \label{embeddings_plot}
\end{figure}

\subsection{Investigating methods for inventing pseudo-subclasses}
\label{unsupervised_clustering_comparison}

In this section we investigate the extent to which inventing pseudo-subclasses via unsupervised clustering of the teacher model’s embeddings can be beneficial for distillation.  Our investigation focuses on tasks with inherent subclass structure, driven by the hypothesis that class embeddings contain rich and informative structural details. Specifically, we utilize binarized versions of CIFAR-10 and CIFAR-100, assigning binary labels based on the original class labels ($y_{binary} = y_{original}\%2$). These datasets present a unique challenge for knowledge distillation, as vanilla distillation is known to underperform due to limited class label information.

Crucially, the availability of original labels ($y_{original}$) in these subclass datasets allows us to explore the "Oracle Clustering" approach. This entails training the student model on the full 10/100-way classification task for CIFAR-10/100, respectively, and subsequently treating the problem as binary classification during testing. Interestingly, the student models trained utilizing the Oracle Clustering approach exhibit the highest performance among all evaluated clustering methods.  This approach not only outperforms other clustering strategies but also surpasses the performance of the original teacher network in certain cases.  A t-SNE visualization of the feature embeddings learned by the student models, presented in Figure~\ref{embeddings_plot}, offers a deeper understanding of the underlying factors contributing to this observed performance advantage.  While Oracle Clustering represents an idealized scenario where subclass structure is known a priori and thus impractical for real-world datasets, its consideration underscores the potential power of inventing pseudo-subclasses as a methodological approach.

We consider several natural ways of inventing pseudo-subclasses by clustering the teacher's embedding space and we compare them with LELP. In particular,  we consider three approaches: (i) Agglomerative clustering (complete linkage); (ii) K-means clustering; and (iii) K-means clustering after we first have projected the teacher's embedding space in a two-dimensional space using t- SNE~\cite{hinton2002stochastic} as proposed in~\cite{yang2023neurons}. For these methods, we tested both hard-label (one-hot vector) and soft-label (weighted by teacher probabilities) representations of the subclasses during student training. We observed that soft-labels, even with temperature tuning, did not significantly impact performance. Therefore, we present results using only hard-label (one-hot vector) subclass representations, which can be found in Table~\ref{tab:clusterings-comparison}  (see also Figure~\ref{fig:clustering_methods}). 

Analysis of Table \ref{tab:clusterings-comparison}  reveals several key findings: (i) LELP consistently outperforms all other clustering methods as well as Vanilla KD across all experimental scenarios; (ii) t-SNE \& K-means generally demonstrate superior performance to Vanilla KD; (iii) Agglomerative clustering and K-means clustering exhibit more varied results in comparison to Vanilla KD. Thus, our findings suggest that the generation of pseudo-subclasses via clustering methodologies holds significant potential, however, the efficacy of this approach is contingent upon the specific clustering algorithm employed.

\begin{table*}[h] 
\renewcommand{\arraystretch}{1.15}
\caption{Comparison of different unsupervised clustering approaches for inventing pseudo-subclasses.}
\label{tab:clusterings-comparison}

\begin{center}
\tiny

\resizebox{1.0\textwidth}{!}{%
\begin{tabular}{|c| c c c c c c |} 
 \hline
 %\rowcolor{cyan}
\cellcolor{cyan}
Teacher Architecture & ResNet92 & ResNet92 & MobileNetWD2 & ResNet92  & ResNet92 & MobileNetWD2\\ 

 %\hline
 % \rowcolor{cyan}
 \cellcolor{cyan}
Student Architecture  & ResNet56 &  MobileNet  & MobileNet& ResNet56  & MobileNet & MobileNet \\ 
%\hline
 \cellcolor{cyan}
Dataset&

CIFAR-10-bin&  
CIFAR-10-bin& 
CIFAR-10-bin& 
CIFAR-100-bin& 
CIFAR-100-bin& 
CIFAR-100-bin \\ 
\hline
 \cellcolor[HTML]{FBB982}
 Teacher & 
 $ 96.70$ &  
 $96.70$ & 
  $93.90 $ &
 $ 78.69 $ & 
 $ 78.69$ &
  $ 72.63$ 
 \\
%\hline
 \cellcolor[HTML]{FBB982}
Oracle Clustering & 
 $ 97.17 \pm 0.02$ &  
 $93.23 \pm 0.28 $ & 
 $93.48 \pm 0.05 $ &
 $ 86.42 \pm 0.11 $ & 
 $ 76.59 \pm 0.60$ &
 $ 77.03 \pm 0.1$ \\

\hline
 \cellcolor[HTML]{FDBC42}
 Vanilla Distillation  &  
 $95.75 \pm 0.21$ & 
 $92.91 \pm 0.17$ & 
 $\underline{92.79 \pm 0.3 }$ &
 $72.87 \pm 0.27$ & 
 $72.16 \pm 0.39$ &
 $\underline{71.45 \pm 0.59}$\\
% \hline

 \cellcolor[HTML]{FDBC42}
 Agglomerative  &  
 $95.48 \pm 0.08$ & 
 $92.16 \pm 0.15$ & 
 $92.41 \pm 0.28 $ &
 $76.84 \pm 0.38$ & 
 $68.05 \pm 1.80$ &
 $66.97 \pm 0.93$\\
% \hline
 \cellcolor[HTML]{FDBC42}
K-means  &  
 $95.76 \pm 0.09$ & 
 $92.53 \pm 0.32$ & 
 $92.05 \pm 0.14 $ &
 $77.03 \pm 0.21$ & 
 $69.4 \pm 0.87$ &
 $68.97 \pm 0.79 $\\
% \hline
 \cellcolor[HTML]{FDBC42}
t-SNE \& K-means \cite{yang2023neurons} &  
 $\underline{95.92 \pm 0.19}$ & 
 $\underline{92.97 \pm 0.07}$ & 
 $91.20 \pm 0.52$ &
 $\underline{77.39 \pm 0.29}$ & 
 $\underline{72.19 \pm 0.60}$ & 
 $68.15 \pm 0.09$\\
%\hline
\cellcolor[HTML]{FDBC42}
LELP (Ours) & 
 $\mathbf{96.71 \pm 0.04}$ & 
 $\mathbf{93.99 \pm 0.17}$ & 
 $ \mathbf{93.03 \pm 0.10}$ &
 $\mathbf{79.91 \pm 0.15}$ & 
 $\mathbf{75.21 \pm 0.12}$ &
 $\mathbf{72.38 \pm 0.24} $\\
 \hline
  
\end{tabular}
}
\end{center}

\end{table*}

\subsection{Comparison with previous approaches} 
\label{comparison_with_baselines}

In this section we compare LELP with other distillation baselines.

\subsubsection{Warmup: Binary Classification with Subclass Structure}
\label{sec:binary_subclass}

 As a warmup,  we first focus on on binary datasets that have an inherent  subclass structure following the setting of Section~\ref{unsupervised_clustering_comparison}. The results can be found in Table~\ref{tab:binary-subclass-structure} in Appendix~\ref{app:binary_subclass_structure_results}.  Among the baselines, Subclass Distillation often performs the best, consistent with the finding that subclass-splitting is effective with little label information~\cite{muller2020subclass}. Our method is able to capture the learned structure from the teacher subclasses without retraining the more expensive teacher model and outperforms all baselines. It's worth noting that our method achieves the highest average improvement over the leading baseline when applied to the CIFAR-100bin dataset, specifically when transferring knowledge from a ResNet teacher model to a MobileNet student model. This scenario is particularly challenging due to the presence of numerous subclasses and significant differences between the teacher and student models in terms of both embedding dimensions and architectures. Interestingly, our method even outperforms Oracle Clustering in the CIFAR-10bin ResNet to MobileNet case, suggesting that it leverages structure that is learned by the teacher that is not present in the original CIFAR-10 labels.

\subsubsection{Few-Class Classification without Subclass Structure}
\label{NLP_classification}

In this section we focus on datasets without subclass structure. We consider classification on six  language classification tasks: the Large Movie Review dataset \cite{maas-EtAl:2011:ACL-HLT2011}, two GLUE datasets: cola and sst2 \cite{wang2018glue}, two datasets sampled from the Amazon US reviews datasets~\cite{tensorflow_amazon_us_reviews}, and a dataset sampled from the Sentiment140 dataset~\cite{go2009twitter}. Details are given in Appendix~\ref{app:NLP_datasets_details}. Our results are shown in Table~\ref{tab:nlp_binary}.  Notably, Subclass Distillation is typically the best performing baseline. LELP exceeds or performs as well as Subclass Distillation, but importantly does not require retraining the teacher, which in the age of ever growing large language models, becomes increasingly important. It is also worth noting that for the case of Amazon US Reviews-based datasets (among the largest ones considered in our study) where we distill to an ALBERT-Base model, LELP significantly outperforms even the teacher model, which contains over 20x the number of parameters. 

% Additionally, in Appendix~\ref{app:addtional_t5} we present additional experiments where the students are MLPs operating over T5 (11B)~\citep{T5_sentence} embeddings.

\section{Limitations}
\label{app:limitations}

While our method is simple and efficient to implement, there may be limitations in using a simple linear projection on the teacher final layer embeddings to extract subclass data. Firstly, there is no reason to assume that the subclasses should be linearly separable in the teacher embedding, and it is likely that more sophisticated unsupervised clustering methods could extract richer information. Second, LELP performs when there is limited teacher logit information (such as in binary classification tasks), however larger image datasets with many classes  contain sufficient information in their teacher logits, obviating the need of subclass splitting methods such as LELP. Indeed, as the number of classes increases, we anticipate LELP’s performance to converge with vanilla knowledge distillation. (Consequently, we did not  present experiments on
datasets with a large number of classes, such as ImageNet-1k, since LELP is not designed for such scenarios.)

\section{Conclusion}

In this study, we have presented evidence that the creation of pseudo-subclasses via unsupervised clustering of teacher embeddings can improve distillation performance in few-class classification tasks, without necessitating the retraining of the teacher model. Through empirical evaluation, we observed that linear projections consistently yield high performance, prompting us to introduce LELP. The superior performance of the "Oracle Clustering" method, where subclass structure is known a priori, suggests that the generation of pseudo-subclasses through clustering techniques has substantial promise. Consequently, future research can explore more sophisticated methods for extracting teacher embedding information, drawing insights from the expanding body of work on Neural Collapse, or investigate strategies for distilling intermediate layer embeddings using a similar approach.

\definecolor{beige}{RGB}{245,245,220}

\begin{table*}[h] 
\centering
\renewcommand{\arraystretch}{1.15}
\caption{Experiments on  Classification Tasks \textbf{without} Subclass Structure.}
\label{tab:nlp_binary}

\begin{center}
\tiny
\resizebox{0.9\textwidth}{!}{%
\begin{tabular}{|c| c c c c c c c c|} 
 \hline
 %\rowcolor{cyan}
\cellcolor{cyan}
Teacher Architecture & 
ALBERT-Large & 
ALBERT-Large  &
ALBERT-Large & 
ALBERT-XXL & 
ALBERT-XXL & 
ALBERT-XXL &
ALBERT-XXL &
ALBERT-XL \\ 

 %\hline
 % \rowcolor{cyan}
 \cellcolor{cyan}
Student Architecture  & 
ALBERT-Base & 
ALBERT-Base  & 
ALBERT-Base  & 
ALBERT-Base  &
MLP/T5(11B)  &
ALBERT-Base  &
MLP/T5(11B)  &
ALBERT-Base \\ 
%\hline
 \cellcolor{cyan}
Dataset&

LMRD&  
GLUE/cola& 
GLUE/sst2 &
Am. Reviews Bin &
Am. Reviews Bin &
Am. Reviews &
Am. Reviews &
Sentiment140 Bin \\
\hline
 \cellcolor[HTML]{FBB982}
 Teacher & 
 $ 90.19$ &  
 $81.87$ & 
 $ 94.09$ &
 $87.82$ &
 $87.82$ &
 $77.58$ &
 $77.58$ &
 $87.29 $  \\
%\hline
 \cellcolor[HTML]{FBB982}
 Subclass Distillation Teacher & 
 $ 90.05 $ &  
 $81.49 $ & 
 $ 93.11 $ &
 $87.71$ &
 $87.53$ &
$78.45$  &
$78.02$ &
$87.45$\\

\hline
\cellcolor[HTML]{FDBC42}
Standard Training & 
 $ 87.68 \pm 0.46$ &  
 $ 79.13 \pm 0.79 $ & 
 $ 91.32 \pm 0.12$ &
 $85.95 \pm 0.09 $ &
 $87.83 \pm 1.48$ &
$74.51 \pm 0.32$ &
$66.44 \pm 0.23$ &
 $ 85.82 \pm 0.15  $   \\

 %\hline
\cellcolor[HTML]{FDBC42}
 Vanilla Distillation & 
 $ 88.70 \pm 0.17$ &  
 $ 80.50 \pm 0.05 $ & 
 $ 92.39 \pm 0.24$ &
 $86.76 \pm 0.23$ &
 $88.60 \pm 0.1$ &
 $75.13 \pm 0.12 $ &
 $72.30 \pm 1.19$ &
  $ 85.95 \pm 0.17$   \\

 %\hline
 \cellcolor[HTML]{FDBC42}
 Embedding Distillation &   
 $ 88.98 \pm 0.33$ & 
 $ 80.66 \pm 0.65$ & 
 $  92.62 \pm 0.16 $&
 $86.43 \pm 0.14 $ &
 $89.84 \pm 0.78$ &
 $75.28 \pm 0.07$  &
 $72.64 \pm 0.95$ &
 $ 85.78 \pm 0.03 $   \\
 
 %\hline
 \cellcolor[HTML]{FDBC42}
 FitNet &   
 $ 88.75 \pm 0.15$ & 
 $ 80.82 \pm 0.32$ & 
 $ 92.43 \pm 0.18$ &
 $86.53 \pm 0.2 $ &
 $\underline{91.40 \pm 0.5}$ &
 $75.54 \pm 0.05$ & 
 $75.78 \pm 0.21$ &
 $86.07 \pm 0.02 $\\

 %\hline
 \cellcolor[HTML]{FDBC42}
 VID  & 
 $88.26 \pm 0.20$ & 
 $79.83 \pm 0.62$ & 
 $91.47 \pm 0.33$ & 
 $83.74 \pm 0.7 $ &
 $89.23 \pm 0.91$ &
$67.48 \pm 0.81 $ &
$63.98 \pm 1.79 $ &
$ 85.83 \pm 0.01 $  \\
 %\hline
 \cellcolor[HTML]{FDBC42}
 Relational KD  & 
 $88.90 \pm 0.24$ & 
 $\underline{81.24 \pm 0.43}$ & 
 $92.73 \pm 0.15$ & 
 $86.36 \pm 0.09$ &
 $89.52 \pm 0.95$ &
$75.12 \pm 0.39 $ &
$74.72 \pm 0.91$ &
$\underline{86.12 \pm 0.07}$\\

 \cellcolor[HTML]{FDBC42}
 DKD  & 
 $88.70 \pm 0.17$ & 
 $80.50 \pm 0.05$ & 
 $92.39 \pm 0.24$  & 
 $86.76 \pm 0.23$  &
 $88.6 \pm 0.1$ &
$72.41 \pm 0.91 $ & 
$72.40 \pm 0.82$ &
 $85.95 \pm 0.17 $ \\

  \cellcolor[HTML]{FDBC42}
 CRD  & 
 $89.19 \pm 0.03$ & 
 $80.79 \pm 0.21$ & 
 $92.27 \pm 0.48$ & 
 $85.79 \pm 0.11$ &
 $88.12 \pm 0.91$ &
$74.7 \pm 1.32 $ & 
$61.08 \pm 1.8$ &
$84.60 \pm 0.69$\\

 \cellcolor[HTML]{FDBC42}
 Subclass Distillation  &  
 $\mathbf{89.24 \pm 0.31}$ & 
 $80.85 \pm 0.1$ & 
 $\mathbf{92.85 \pm 0.15}$ &
 $\underline{87.34 \pm 0.1}$ &
 $ 90.38 \pm 0.82 $ &
$\underline{76.23 \pm 0.50}$ & 
 $\underline{77.27 \pm 1.45 }$ &
$ 85.93 \pm 0.24 $\\

 %\hline
 \cellcolor[HTML]{FDBC42}
LELP (Ours) &  
 $\underline{89.22 \pm 0.06}$ & 
 $\mathbf{81.43 \pm 0.47}$ & 
 $\underline{92.81 \pm 0.36}$ &
$\mathbf{88.49 \pm 0.36}$ &
$\mathbf{91.76 \pm 0.17}$ &
$\mathbf{78.08 \pm 0.81}$ &
$\mathbf{77.48 \pm 0.43 } $ &
$ \mathbf{87.00 \pm 0.25} $\\
 
 \hline
  \cellcolor{lightgray}
 Avg. gain over the best baseline&
\cellcolor{beige} $-0.02$ & 
\cellcolor{green} $0.20$ & 
\cellcolor{beige} $-0.04$&
\cellcolor{green} $1.15$ & 
\cellcolor{green} $0.36$ & 
\cellcolor{green} $1.85$ &
\cellcolor{green} $0.21$ &
\cellcolor{green}  $0.88$  \\ 

% \hline
  \cellcolor{lightgray}
 Avg. gain over non-subclass baseline &
\cellcolor{green} $0.24$ & 
\cellcolor{green} $0.20$ & 
\cellcolor{green} $0.08$ &
\cellcolor{green} $1.73$ & 
\cellcolor{green} $0.36$ & 
\cellcolor{green} $2.54$ &
\cellcolor{green} $1.68$ & 
\cellcolor{green} $0.88$ \\ 

 % \hline
\cellcolor{lightgray}
 Avg. gain over Vanilla KD&
\cellcolor{green} $0.52$ & 
\cellcolor{green} $0.93$ & 
\cellcolor{green} $0.42$ &
\cellcolor{green} $1.73$ &
\cellcolor{green} $3.16$ &
\cellcolor{green} $2.95$ &
\cellcolor{green} $5.18$ &
\cellcolor{green} $1.18$ \\

\hline
  
\end{tabular}
}
\end{center}
\vskip -0.2in

\end{table*}

\bibliographystyle{plain}
\bibliography{references}

\begin{thebibliography}{10}

\bibitem{abadi2016tensorflow}
Mart{\'\i}n Abadi, Ashish Agarwal, Paul Barham, Eugene Brevdo, Zhifeng Chen,
  Craig Citro, Greg~S Corrado, Andy Davis, Jeffrey Dean, Matthieu Devin, et~al.
\newblock Tensorflow: Large-scale machine learning on heterogeneous distributed
  systems.
\newblock {\em arXiv preprint arXiv:1603.04467}, 2016.

\bibitem{Ahn2019CVPR}
Sungsoo Ahn, Shell~Xu Hu, Andreas Damianou, Neil~D. Lawrence, and Zhenwen Dai.
\newblock Variational information distillation for knowledge transfer.
\newblock In {\em Proceedings of the IEEE/CVF Conference on Computer Vision and
  Pattern Recognition (CVPR)}, June 2019.

\bibitem{arazo2020pseudo}
Eric Arazo, Diego Ortego, Paul Albert, Noel~E O’Connor, and Kevin McGuinness.
\newblock Pseudo-labeling and confirmation bias in deep semi-supervised
  learning.
\newblock In {\em 2020 International Joint Conference on Neural Networks
  (IJCNN)}, pages 1--8. IEEE, 2020.

\bibitem{RAD}
Cenk Baykal, Khoa Trinh, Fotis Iliopoulos, Gaurav Menghani, and Erik Vee.
\newblock Robust active distillation.
\newblock {\em International Conference on Learning Representations (ICLR)},
  2023.

\bibitem{brown2020language}
Tom Brown, Benjamin Mann, Nick Ryder, Melanie Subbiah, Jared~D Kaplan, Prafulla
  Dhariwal, Arvind Neelakantan, Pranav Shyam, Girish Sastry, Amanda Askell,
  et~al.
\newblock Language models are few-shot learners.
\newblock {\em Advances in neural information processing systems},
  33:1877--1901, 2020.

\bibitem{bucilua}
Cristian Buciluǎ, Rich Caruana, and Alexandru Niculescu-Mizil.
\newblock Model compression.
\newblock In {\em Proceedings of the 12th ACM SIGKDD international conference
  on Knowledge discovery and data mining}, pages 535--541, 2006.

\bibitem{chen2020big}
Ting Chen, Simon Kornblith, Kevin Swersky, Mohammad Norouzi, and Geoffrey~E
  Hinton.
\newblock Big self-supervised models are strong semi-supervised learners.
\newblock {\em Advances in neural information processing systems},
  33:22243--22255, 2020.

\bibitem{tensorflow_amazon_us_reviews}
TensorFlow Datasets.
\newblock Amazon us reviews dataset, 2020.

\bibitem{devlin2018bert}
Jacob Devlin, Ming-Wei Chang, Kenton Lee, and Kristina Toutanova.
\newblock Bert: Pre-training of deep bidirectional transformers for language
  understanding.
\newblock {\em arXiv preprint arXiv:1810.04805}, 2018.

\bibitem{fang2021exploring}
Cong Fang, Hangfeng He, Qi~Long, and Weijie~J Su.
\newblock Exploring deep neural networks via layer-peeled model: Minority
  collapse in imbalanced training.
\newblock {\em Proceedings of the National Academy of Sciences},
  118(43):e2103091118, 2021.

\bibitem{go2009twitter}
Alec Go, Richa Bhayani, and Lei Huang.
\newblock Twitter sentiment classification using distant supervision.
\newblock {\em CS224N project report, Stanford}, 1(12):2009, 2009.

\bibitem{he2016deep}
Kaiming He, Xiangyu Zhang, Shaoqing Ren, and Jian Sun.
\newblock Deep residual learning for image recognition.
\newblock In {\em Proceedings of the IEEE conference on computer vision and
  pattern recognition}, pages 770--778, 2016.

\bibitem{hinton2002stochastic}
Geoffrey~E Hinton and Sam Roweis.
\newblock Stochastic neighbor embedding.
\newblock {\em Advances in neural information processing systems}, 15, 2002.

\bibitem{distillation}
Geoffrey~E. Hinton, Oriol Vinyals, and Jeffrey Dean.
\newblock Distilling the knowledge in a neural network.
\newblock {\em ArXiv}, abs/1503.02531, 2015.

\bibitem{IKBMTV22}
Fotis Iliopoulos, Vasilis Kontonis, Cenk Baykal, Gaurav Menghani, Khoa Trinh,
  and Erik Vee.
\newblock Weighted distillation with unlabeled examples.
\newblock In {\em NeurIPS}, 2022.

\bibitem{kim2018paraphrasing}
Jangho Kim, SeongUk Park, and Nojun Kwak.
\newblock Paraphrasing complex network: Network compression via factor
  transfer.
\newblock {\em Advances in neural information processing systems}, 31, 2018.

\bibitem{kontonis2023slam}
Vasilis Kontonis, Fotis Iliopoulos, Khoa Trinh, Cenk Baykal, Gaurav Menghani,
  and Erik Vee.
\newblock Slam: Student-label mixing for distillation with unlabeled examples.
\newblock In {\em Thirty-seventh Conference on Neural Information Processing
  Systems}, 2023.

\bibitem{krizhevsky2009learning}
Alex Krizhevsky, Geoffrey Hinton, et~al.
\newblock Learning multiple layers of features from tiny images.
\newblock 2009.

\bibitem{liu2021certainty}
Lu~Liu and Robby~T Tan.
\newblock Certainty driven consistency loss on multi-teacher networks for
  semi-supervised learning.
\newblock {\em Pattern Recognition}, 120:108140, 2021.

\bibitem{maas-EtAl:2011:ACL-HLT2011}
Andrew~L. Maas, Raymond~E. Daly, Peter~T. Pham, Dan Huang, Andrew~Y. Ng, and
  Christopher Potts.
\newblock Learning word vectors for sentiment analysis.
\newblock In {\em Proceedings of the 49th Annual Meeting of the Association for
  Computational Linguistics: Human Language Technologies}, pages 142--150,
  Portland, Oregon, USA, June 2011. Association for Computational Linguistics.

\bibitem{menghani2021efficient}
G~Menghani.
\newblock Efficient deep learning: A survey on making deep learning models
  smaller.
\newblock {\em Faster, and Better. arXiv}, 2106, 2021.

\bibitem{muller2020subclass}
Rafael M{\"u}ller, Simon Kornblith, and Geoffrey Hinton.
\newblock Subclass distillation.
\newblock {\em arXiv preprint arXiv:2002.03936}, 2020.

\bibitem{T5_sentence}
Jianmo Ni, Gustavo~Hern{\'{a}}ndez {\'{A}}brego, Noah Constant, Ji~Ma, Keith~B.
  Hall, Daniel Cer, and Yinfei Yang.
\newblock Sentence-t5: Scalable sentence encoders from pre-trained text-to-text
  models.
\newblock {\em CoRR}, abs/2108.08877, 2021.

\bibitem{papyan2020prevalence}
Vardan Papyan, XY~Han, and David~L Donoho.
\newblock Prevalence of neural collapse during the terminal phase of deep
  learning training.
\newblock {\em Proceedings of the National Academy of Sciences},
  117(40):24652--24663, 2020.

\bibitem{park2019relational}
Wonpyo Park, Dongju Kim, Yan Lu, and Minsu Cho.
\newblock Relational knowledge distillation.
\newblock In {\em Proceedings of the IEEE/CVF conference on computer vision and
  pattern recognition}, pages 3967--3976, 2019.

\bibitem{passalis2018unsupervised}
Nikolaos Passalis and Anastasios Tefas.
\newblock Unsupervised knowledge transfer using similarity embeddings.
\newblock {\em IEEE transactions on neural networks and learning systems},
  30(3):946--950, 2018.

\bibitem{pham2021meta}
Hieu Pham, Zihang Dai, Qizhe Xie, and Quoc~V Le.
\newblock Meta pseudo labels.
\newblock In {\em Proceedings of the IEEE/CVF Conference on Computer Vision and
  Pattern Recognition}, pages 11557--11568, 2021.

\bibitem{romero2014fitnets}
Adriana Romero, Nicolas Ballas, Samira~Ebrahimi Kahou, Antoine Chassang, Carlo
  Gatta, and Yoshua Bengio.
\newblock Fitnets: Hints for thin deep nets.
\newblock {\em arXiv preprint arXiv:1412.6550}, 2014.

\bibitem{simonyan2014very}
Karen Simonyan and Andrew Zisserman.
\newblock Very deep convolutional networks for large-scale image recognition.
\newblock {\em arXiv preprint arXiv:1409.1556}, 2014.

\bibitem{stanton2021does}
Samuel Stanton, Pavel Izmailov, Polina Kirichenko, Alexander~A Alemi, and
  Andrew~G Wilson.
\newblock Does knowledge distillation really work?
\newblock {\em Advances in Neural Information Processing Systems},
  34:6906--6919, 2021.

\bibitem{tian2019contrastive}
Yonglong Tian, Dilip Krishnan, and Phillip Isola.
\newblock Contrastive representation distillation.
\newblock {\em arXiv preprint arXiv:1910.10699}, 2019.

\bibitem{wang2018glue}
Alex Wang, Amanpreet Singh, Julian Michael, Felix Hill, Omer Levy, and Samuel~R
  Bowman.
\newblock Glue: A multi-task benchmark and analysis platform for natural
  language understanding.
\newblock {\em arXiv preprint arXiv:1804.07461}, 2018.

\bibitem{xie2020self}
Qizhe Xie, Minh-Thang Luong, Eduard Hovy, and Quoc~V Le.
\newblock Self-training with noisy student improves imagenet classification.
\newblock In {\em Proceedings of the IEEE/CVF conference on computer vision and
  pattern recognition}, pages 10687--10698, 2020.

\bibitem{yang2023neurons}
Yongyi Yang, Jacob Steinhardt, and Wei Hu.
\newblock Are neurons actually collapsed? on the fine-grained structure in
  neural representations.
\newblock In {\em International Conference on Machine Learning}, pages
  39453--39487. PMLR, 2023.

\bibitem{zhao2022decoupled}
Borui Zhao, Quan Cui, Renjie Song, Yiyu Qiu, and Jiajun Liang.
\newblock Decoupled knowledge distillation.
\newblock In {\em Proceedings of the IEEE/CVF Conference on computer vision and
  pattern recognition}, pages 11953--11962, 2022.

\end{thebibliography}
\newpage
\appendix
\onecolumn

\section{Broader Impact Statement}
\label{app:broader_impact}
Due to its popularity and the fact that deep learning is widely used in fields from NLP to robotics to autonomous vehicles, knowledge distillation, as a deep learning method, carries with it the potential for harmful societal impact. However, we feel that none of these impacts must be specifically highlighted here.

\section{Binary Classification with Subclass Structure: Experimental Results}
\label{app:binary_subclass_structure_results}

Here we present the table with the experimental results corresponding to Sections~\ref{unsupervised_clustering_comparison} and~\ref{sec:binary_subclass}.

\begin{table*}[ht!] 
\centering
\renewcommand{\arraystretch}{1.15}
\caption{Experiments on Binary Classification tasks \textbf{with} Subclass Structure.}
\label{tab:binary-subclass-structure}
\begin{center}
\tiny
\resizebox{1.0\textwidth}{!}{%
\begin{tabular}{|c| c c c c c c |} 
 \hline
 %\rowcolor{cyan}
\cellcolor{cyan}
Teacher Architecture & ResNet92 & ResNet92  & MobileNetWD2 & ResNet92  & ResNet92 & MobileNetWD2 \\ 

 %\hline
 % \rowcolor{cyan}
 \cellcolor{cyan}
Student Architecture  & ResNet56 &  MobileNet  & MobileNet &  ResNet56  & MobileNet & MobileNet \\ 
%\hline
 \cellcolor{cyan}
Dataset&

CIFAR-10-bin&  
CIFAR-10-bin& 
CIFAR-10-bin& 
CIFAR-100-bin& 
CIFAR-100-bin& 
CIFAR-100-bin \\ 
\hline
 \cellcolor[HTML]{FBB982}
 Teacher & 
 $ 96.70$ &  
 $96.70$ & 
  $93.90 $ &
 $ 78.69 $ & 
 $ 78.69$ &
  $ 72.63$ 
 \\
%\hline
 \cellcolor[HTML]{FBB982}
 Subclass Distillation Teacher
 & 
 $ 96.12 $ &  
 $95.99 $ & 
 $93.82$ &
 $ 77.52 $ & 
 $ 75.87 $ &
 $71.53 $ \\
%\hline
 \cellcolor[HTML]{FBB982}
Oracle Clustering & 
 $ 97.17 \pm 0.02$ &  
 $93.48 \pm 0.05 $ & 
 $93.48 \pm 0.05 $ &
 $ 86.59 \pm 0.05 $ & 
 $ 86.38 \pm 0.06$ &
 $ 86.38 \pm 0.06$
 \\
\hline
\cellcolor[HTML]{FDBC42}
Standard Training & 
 $ 95.43 \pm 0.07$ &  
 $ 92.04 \pm 0.26 $ & 
 $ 92.13  \pm 0.17 $ &
 $ 74.80 \pm 0.52 $ & 
 $ 69.13 \pm 0.32$ &
 $ 68.84 \pm 0.97 $ \\
 
 %\hline
\cellcolor[HTML]{FDBC42}
 Vanilla Distillation& 
 $ 95.75 \pm 0.21$ &  
 $ 92.91 \pm 0.17 $ &
  $ 92.79 \pm 0.3 $ &
 $ 72.87 \pm 0.27 $ & 
 $ 72.16 \pm 0.39$ &
$ 71.45 \pm 0.59 $ \\
 
% \hline
 \cellcolor[HTML]{FDBC42}
 Embedding Distillation &   
 $96.48 \pm 0.1$  & 
 $93.05 \pm 0.13$ & 
 $ 92.57 \pm 0.06 $&
 $78.77 \pm 0.35$ & 
 $72.24 \pm 0.20$ &
$ \underline{71.99 \pm 0.30 }$
 \\
 
% \hline
 \cellcolor[HTML]{FDBC42}
 FitNet&   
 $ \underline{96.53 \pm 0.05}$ & 
 $ 93.31 \pm 0.11$ & 
 $ 92.72 \pm 0.1 $ &
 $ 79.14 \pm 0.18$ & 
 $  71.76 \pm 0.26 $ &
 $ 68.79 \pm 2.82$
 \\
 %\hline
 \cellcolor[HTML]{FDBC42}
 VID  & 
 $96.13 \pm 0.16$ & 
 $89.95 \pm 1.34$ & 
 $84.09 \pm 3.38$ &
 $75.69 \pm 0.55$ & 
 $71.99 \pm 0.18$ & 
 $61.15 \pm 0.40 $\\  
% \hline
 \cellcolor[HTML]{FDBC42}
  Relational KD  & 
 $96.22 \pm 0.09$ & 
 $92.98 \pm 0.21$ & 
 $\underline{92.81 \pm 0.14}$ &
 $78.57 \pm 0.64$ & 
 $72.53 \pm 0.23$ & 
 $71.66 \pm 0.1$\\  
% \hline
 \cellcolor[HTML]{FDBC42}
  DKD  & 
  $95.75 \pm 0.21$ & 
  $92.91 \pm 0.17$ & 
  $92.79 \pm 0.3$ &
  $72.87 \pm 0.27$ & 
  $72.16 \pm 0.39$ & 
  $71.45 \pm 0.59$  \\  
 \cellcolor[HTML]{FDBC42}
  CRD  & 
 $ 95.83 \pm 0.25$ & 
 $92.47 \pm 0.13$ & 
 $ 92.34 \pm 0.18$ &
  $73.01 \pm 0.21 $ & 
 $72.15 \pm 0.63$ & 
 $69.53 \pm  1.37$ \\

 \cellcolor[HTML]{FDBC42}
 Subclass Distillation  &  
 $96.44 \pm 0.06$ & 
 $\underline{93.76 \pm 0.06}$ & 
  $ 92.77 \pm 0.15$  &
 $\underline{79.23 \pm 0.17}$ & 

 $\underline{73.90 \pm 0.18}$ &
 $70.60 \pm 0.27$
 \\
  \cellcolor[HTML]{FDBC42}
 Agglomerative  &  
 $95.48 \pm 0.02$ & 
 $92.16 \pm 0.15$ & 
 $92.41 \pm 0.28$ &
 $76.84 \pm 0.38$ & 
 $ 68.05 \pm 1.80$ &
 $66.97 \pm 0.93$ 
 \\
\cellcolor[HTML]{FDBC42}
 K-means  &  
 $95.76 \pm 0.09$ & 
 $92.53 \pm 0.32 $ & 
 $92.05 \pm 0.14 $ &
 $77.03 \pm 0.21 $ & 
 $69.4 \pm 0.87 $ &
 $68.97 \pm 0.09 $ 
 \\
 \cellcolor[HTML]{FDBC42}
 t-SNE \& K-means &  
 $95.92 \pm 0.19$ & 
 $92.97 \pm 0.07$ & 
 $91.20 \pm 0.52 $ &
 $77.39 \pm 0.29$ & 
 $72.19 \pm 0.60 $ &
 $68.15 \pm 0.09$ 
 \\
% \hline
 \cellcolor[HTML]{FDBC42}
LELP (Ours) &  
 $\mathbf{96.71 \pm 0.04}$ & 
 $\mathbf{93.99 \pm 0.17}$ & 
 $ \mathbf{93.03 \pm 0.1 }$ &
 $\mathbf{79.91 \pm 0.15}$ & 
 $\mathbf{75.21 \pm 0.12}$ & 
 $ \mathbf{72.38 \pm 0.24}$\\
 \hline
  \cellcolor{lightgray}
 Avg. gain over the best baseline &
\cellcolor{green} $0.23$ & 
\cellcolor{green} $0.23$ &
\cellcolor{green} $0.22$ &
\cellcolor{green} $0.68$ & 
\cellcolor{green} $1.31$ &
\cellcolor{green} $0.39$\\
%  \hline
\cellcolor{lightgray}
 Avg. gain over non-subclass baseline &
\cellcolor{green} $0.23$ & 
\cellcolor{green} $0.68$ & 
\cellcolor{green} $0.22$ &
\cellcolor{green} $0.77$ & 
\cellcolor{green} $2.68$ &
\cellcolor{green} $0.39$ \\
 % \hline
\cellcolor{lightgray}
 Avg. gain over Vanilla KD &
\cellcolor{green} $0.96$ & 
\cellcolor{green} $1.08$ & 
\cellcolor{green} $0.24$ & 
\cellcolor{green} $7.04$ & 
\cellcolor{green} $3.05$ &
\cellcolor{green} $0.93$ \\
  \hline
  
\end{tabular}
}
\end{center}

\vskip -0.2in
\end{table*}

\section{Ablations}
\label{app:ablations}

\begin{figure*}[h!]
\vskip 0.1in
\begin{center}
\includegraphics[width = 1.0\linewidth]{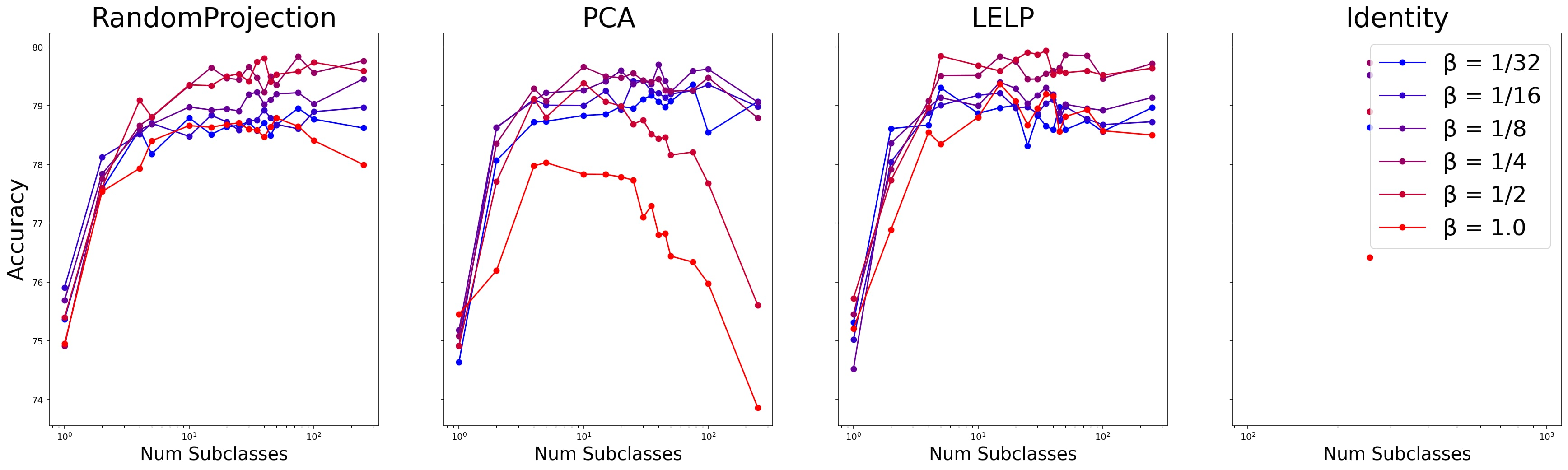}
\includegraphics[width = 1.0\linewidth]{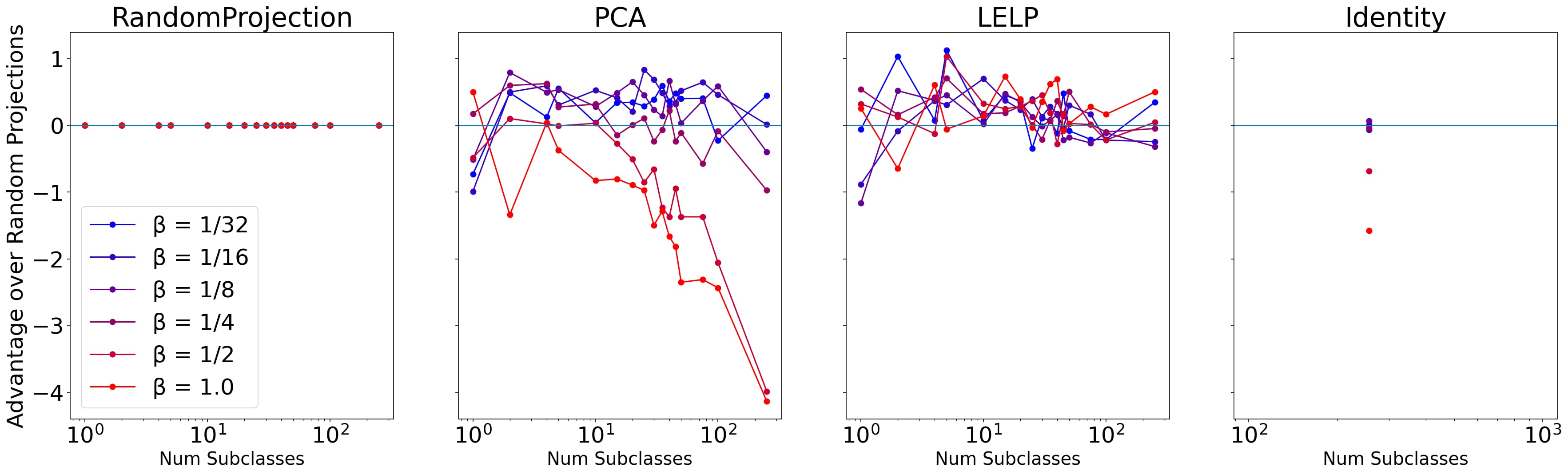}
\caption{\textbf{Top row}: Ablations of choice of Projection, number of subclasses $S$ and subclass tempereature $\beta$ on CIFAR-100bin. The set of plots displays raw CIFAR-100bin accuracy.  \textbf{Bottom row}: The second set of plots demonstrates the accuracy gain achieved over random projections (using the same hyperparameter choice). Values over 0 indicate an advantage over random projections, which we see consistently with LELP.}
\label{fig:ablations_fig}
\end{center}
\vskip -0.2in
\end{figure*}

One of the key design choices in LELP is to use the top PCA directions to construct the pseudo-subclasses. This is based on the intuition that axis which contain the most variation also contain the most information for distillation. Here we test this assumption by comparing PCA projections against three baselines. The first baseline is random projections (\textbf{Rand}), in which we choose $S$ random orthogonal directions to use as our subclass direction. The second baseline is raw PCA, which is using the top-K PCA directions, without the additional projection and random rotation applied, as we discussed in Section~\ref{sec:our_method}. This corresponds to using vectors $v_{c,s}$ instead of $\tilde{v}_{c,s}$. Our final baseline is the ``Identity" projection, which is the same as using all of the teacher embeddings as subclasses. We consider the binarized CIFAR-100 task, and sweep $S$ from $2 - 256$, the embedding dimension of the teacher $D = 256$. We distill from ResNet92 to ResNet56, with the subclass temperature parameter $\beta$ ranging from $2^{-5} - 2^{0}$, with results shown in Figure~\ref{fig:ablations_fig} (top row), averaged over three runs. For a more clear comparison to the random projection baseline, we also show the ``advantange'' over random projections in \cref{fig:ablations_fig} (bottom row), made by subtracting the accuracy by the equivalent obtained by random projects with the same $S$ and $\beta$.  Figure~\ref{fig:ablations_fig} shows the following general trends:

\textbf{Larger $S$ generally improves performance for Random Projections and LELP, and can harm PCA}. For Rand and LELP, performance seems mostly monotonically increasing with larger $S$. For LELP the benefit plateaus around $S = 32$, while for Rand, the performance slowly increases until $S=256$. This suggests that the PCA step used in LELP obtains the salient information from teacher embeddings in fewer $S$ than Rand. For PCA without the extra steps used in LELP, large subclasses with higher temperatures significantly degrades performance, while adding rotation makes the performance of PCA stable. We conjecture that this is due to the first few PCA directions containing most of the variation and the remaining direction behaving as random noise.

\textbf{LELP consistently outperforms Random Projections}. We see in \cref{fig:ablations_fig}, when compared to random projections, LELP almost always has a non-trivial advantage. The performance of Identity projections is more inconsistent and it can be worse than random projections, depending on $\beta$. For larger number of subclasses, the advantage of LELP over random projections diminishes, as both methods now contain the majority of information from the entire embedding space.

\section{Detailed description of the NLP datasets we used.}\label{app:NLP_datasets_details}

The Corpus of Linguistic Acceptability (GLUE/cola) is a binary classification task which consists of English acceptability judgments drawn from books and journal articles on linguistic theory. Each example is a sequence of words annotated with whether it is a grammatical English sentence. GLUE/cola contains $8,551$ examples for training and $1,063$ examples for testing.

The Large Movie Review dataset (LMRD) is a dataset for binary sentiment classification containing $25,000$  movie reviews for training, and $25,000$ for testing.

The Stanford Sentiment Treebank (GLUE/sst2) consists of sentences from movie reviews and human annotations of their sentiment. The task is to predict the sentiment (positive or negative) of a given sentence. It contains $67,349$ examples for training and $1,821$ examples for testing.

The Amazon US reviews datasets comprise a vast collection of over a hundred million customer reviews and ratings (ranging from 1 to 5 stars). For our study, we selected a subset of these reviews across various products and devised two classification tasks:   a 5-classes classification task and a binary classification task, which we describe below.
\begin{itemize}
\item ``Amazon Reviews" contains $500,000$ examples for training and $10,000$ examples for testing. The task is to predict the star rating of the given review (ranging from 1 to 5). Both the training and test sets are balanced in terms of the number of examples per rating.

\item ``Amazon Reviews Bin" is a subset of ``Amazon Reviews'', where we exclude all 3-star reviews. It consists of $400,000$ training  and $8,000$ testing examples. The objective is to determine whether a  review is ``polarized'' (rated either 1 or 5 stars) or ``mild'' (2 or 4 stars).

\end{itemize}

For sampling the examples Amazon Reviews dataset we use the following process. We first first sequentially parse and concatenate the first 200k examples of the following datasets: ``Office Products'', ``Video Games'', ``Video'', ``Toys'', ``Tools'', ``Sports'', ``Jewelry'', ``Digital Music Purchase'', ``Mobile Apps'', ``Video DVD'', ``Watches'' (in this order). We then sequentially parse the resulting collection of examples and we add examples to the training and testing datasets. In particular, for each class (1-5 stars), we add the first 100k examples we encounter to the training dataset, and the next 2k examples we encounter to the test dataset.

The Sentiment140 dataset contains $1,600,000$ tweets extracted using the twitter api, which have been annotated either as positive or negative. We sample $1,590,000$ of these tweets to create a balanced training set, and we use  the rest $10,000$ as the test set (also balanced). Note that the Sentiment140 datasets also comes with a validation set of $498$ examples, part of which is annotated as ``Neutral''. We are not making use of the latter validation set which is why in our tables we refer to the dataset we  use (described earlier) as ``Sentiment140 Bin''.

% \section{Additional NLP experiments:  MLPs over frozen  sentence-T5 (11B) encoder}
% \label{app:addtional_t5}
% \input{tables/t5mlp}

% In this section we present additional NLP experiments.

% Using a pre-trained, frozen large-scale encoder model to generate representations is a common and effective approach when one needs multiple models for different classification tasks. Essentially, this encoder acts as a foundation, with separate, smaller neural networks (MLPs) attached to it for each task. This method is much more efficient in terms of training time and storage space compared to fine-tuning and storing the entire large-scale encoder for each individual task.

% In Table~\ref{tab:nlp_t5} we present experiments in the NLP datasets we have considered, with the student model architecture being a two-layer-MLP of width 4096 operating over representations generated by a pre-trained T5-sentence encoder of 11 Billion parameters \citep{T5_sentence}.  

% \input{tables/crd_dkd}

\section{Multiclass Classification}

\label{app:multiclass}

While our primary focus lies on classification tasks with few classes, here we present results demonstrating the applicability of LELP to multiclass classification tasks with a moderate number of classes. To this end, we evaluate LELP on the CIFAR-10 and CIFAR-100 datasets. It is important to note that our objective is not to establish state-of-the-art performance in vision tasks but rather to showcase the versatility of our approach. We selected CIFAR-10 and CIFAR-100 as our datasets because they are widely recognized benchmarks for 10-class and 100-class classification, respectively.  We intentionally disregarded image-specific techniques, such as data augmentation, to focus on a more general comparison. Consequently, we did not include methods like CRD and DKD that are specifically tailored to image data. Our results indicate that LELP can effectively handle tasks with a moderate number of classes and outperforms modality-independent knowledge distillation methods, particularly in scenarios where there is a significant disparity between the teacher and student architectures. As the number of classes increases, we anticipate LELP’s performance to converge with vanilla knowledge distillation. Consequently, we did not conduct experiments on datasets with a large number of classes, such as ImageNet-1k.

Our results are shown in Table~\ref{tab:multiclass}, where we distill a ResNet92 to a MobileNet.  In this comparison,  LELP significantly outperforms baselines, seeing a 1.1\% improvement and 2.31\% improvement over the next best baseline in CIFAR-10 and CIFAR-100, respectively. This validates the hypothesis that providing subclass information in the form of a classification loss avoids the pitfalls that methods which directly match embeddings have when students and teachers differ. When there is either a large performance or architecture gap, embeddings do not transfer readily. Appendix ~\ref{app:semi} contains additional experiments in semi-supervised knowledge distillation, where we  find that LELP consistently outperforms existing baselines without modification in the semi-supervised setting.

\begin{table*}[ht!] 
\renewcommand{\arraystretch}{1.15}
\caption{Experiments on \textbf{Multi-class Classification tasks}.}
\label{tab:multiclass}

\begin{center}
\tiny
\begin{tabular}{|c| c  c |} 
 \hline
 %\rowcolor{cyan}
\cellcolor{cyan}
Teacher Architecture 
& ResNet92
& ResNet92
\\ 

 %\hline
 % \rowcolor{cyan}
 \cellcolor{cyan}
Student Architecture& 
MobileNet  & 
MobileNet\\ 
% \hline
 \cellcolor{cyan}
Dataset&
CIFAR-10&
CIFAR-100\\ 
\hline
 \cellcolor[HTML]{FBB982}
 Teacher &  
 $94.94$ & 
 $ 75.55$ \\
% \hline
 \cellcolor[HTML]{FBB982}
 Subclass Distillation Teacher
 & 
 $93.49 $ & 
 $ 70.23 $ \\
\hline
\cellcolor[HTML]{FDBC42}
Standard Training &  
 $ 84.86 \pm 0.11 $ &
 $ 53.15 \pm 0.21$ \\
 
 %\hline
\cellcolor[HTML]{FDBC42}
 Vanilla Distillation & 
 $ 86.45 \pm 0.21 $ & 
 $ 56.20 \pm 0.11$ \\
 
% \hline
 \cellcolor[HTML]{FDBC42}
 Embedding Distillation &   
 $86.82 \pm 0.16$ & 
 $57.27 \pm 0.4$ \\
 
 %\hline
 \cellcolor[HTML]{FDBC42}
 FitNet &   
 $ 85.70 \pm 0.36$ & 
 $  38.23 \pm 1.12 $\\
 %\hline
 \cellcolor[HTML]{FDBC42}
 VID  & 
 $76.36 \pm 1.56$ & 
 $27.08 \pm 1.08$ \\  
 %\hline
 \cellcolor[HTML]{FDBC42}
  Relational KD  & 
 $\underline{86.92 \pm 0.34}$ & 

 $\underline{59.67 \pm 0.58}$ \\  
 %\hline
 \cellcolor[HTML]{FDBC42}
 Subclass Distillation &  
 $86.55 \pm 0.28$ & 
 $54.45 \pm 1.25$ \\
 %\hline
 \cellcolor[HTML]{FDBC42}
LELP (Ours) &  
 $\mathbf{88.02 \pm 0.19}$ & 
 $\mathbf{61.98 \pm 0.08}$ \\
 \hline
  \cellcolor{lightgray}
 Avg. gain over the best baseline &
\cellcolor{green} $1.1$ & 
\cellcolor{green} $2.31$ \\
  %\hline
\cellcolor{lightgray}
 Avg. gain over non-subclass baseline &
\cellcolor{green} $1.1$ & 
\cellcolor{green} $2.31$ \\
  %\hline
\cellcolor{lightgray}
 Avg. gain over Vanilla KD &
\cellcolor{green} $1.57$ & 
\cellcolor{green} $5.78$ \\
  \hline
  
\end{tabular}

\end{center}

\vskip -0.2in
\end{table*}

\section{Data Efficiency, Training Speed and Robustness}
\label{app:properties}

\begin{figure*}[h]
\vskip 0.1in
\begin{center}
\includegraphics[width = 0.30\linewidth]{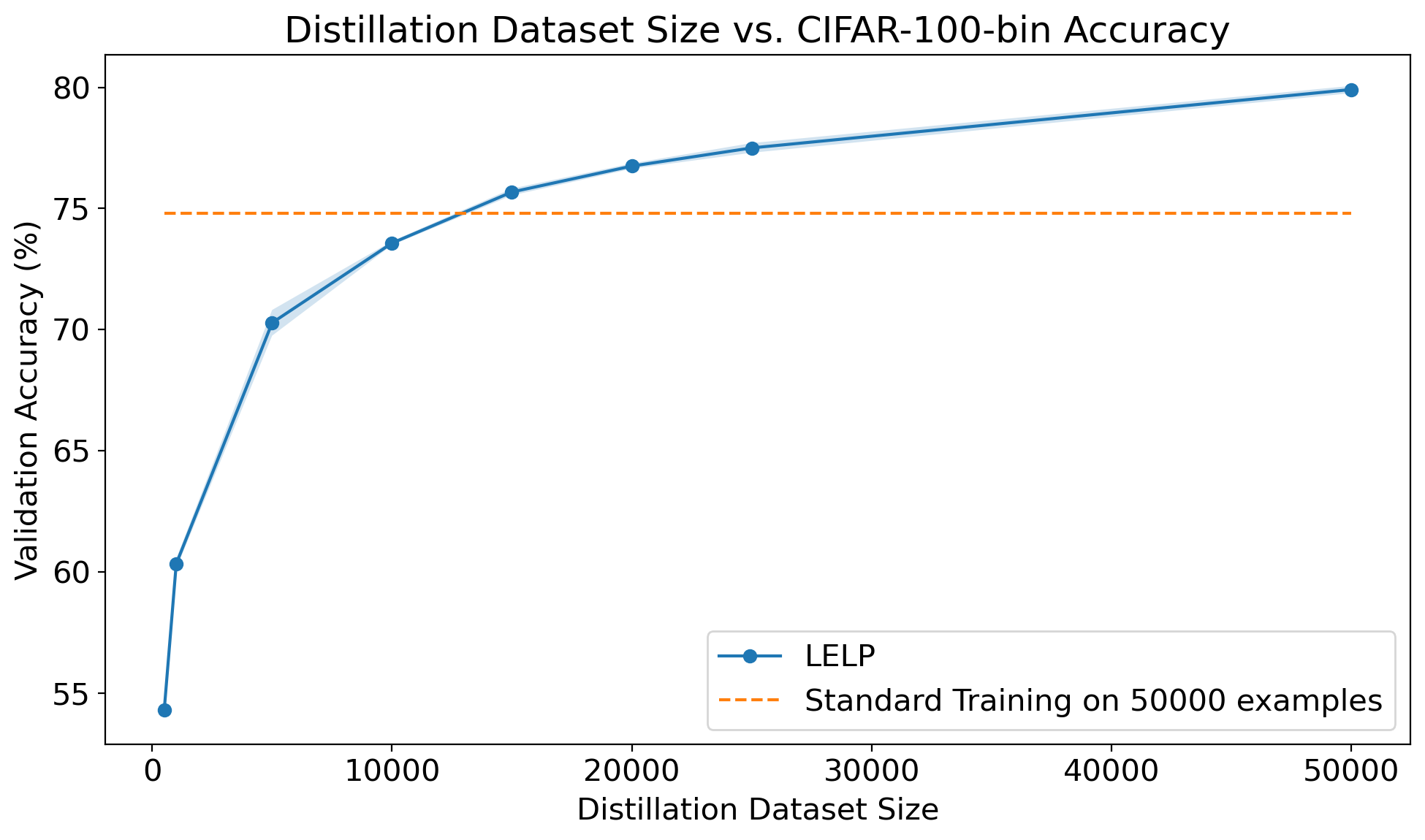}
\includegraphics[width = 0.30\linewidth]{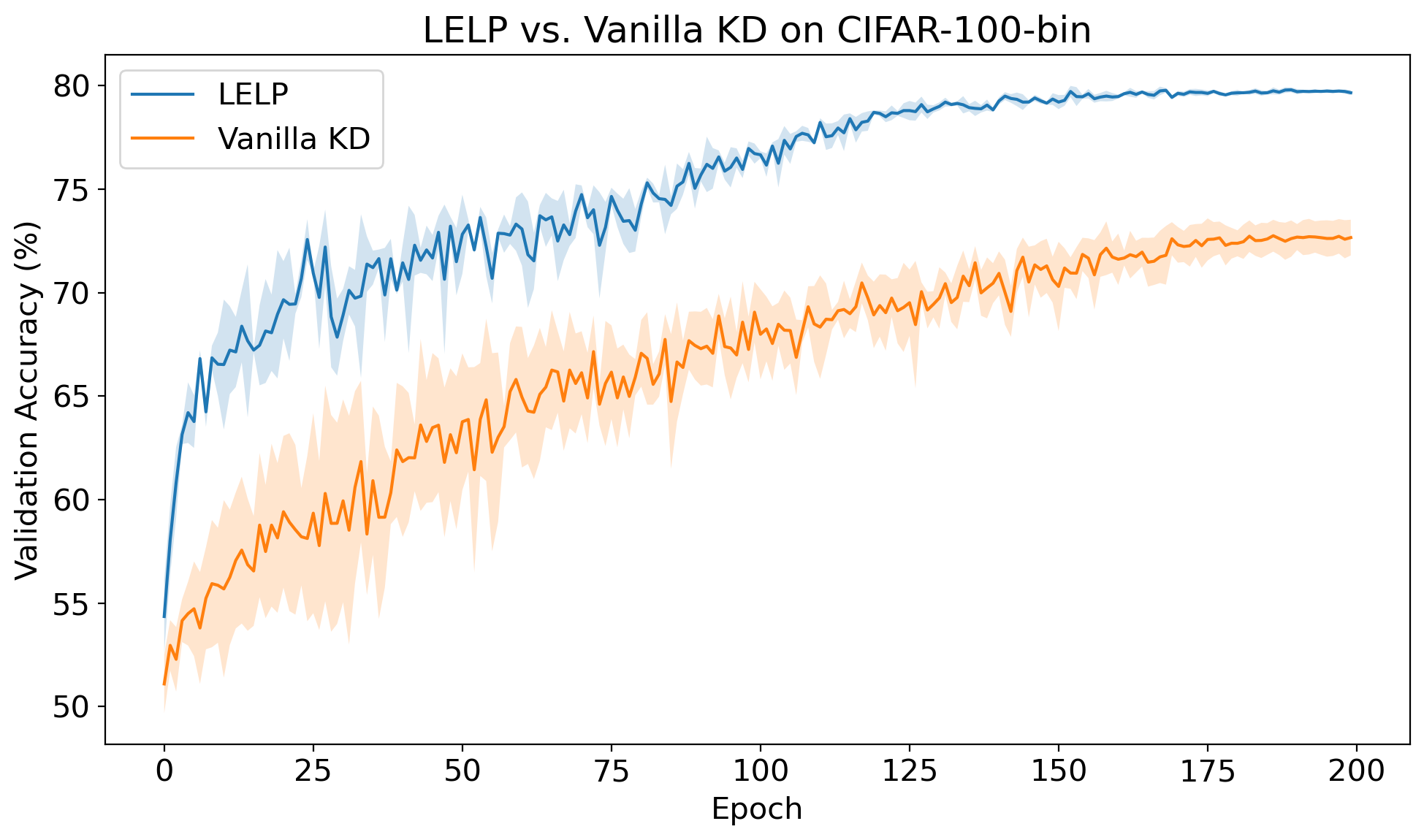}
\includegraphics[width = 0.30\linewidth]{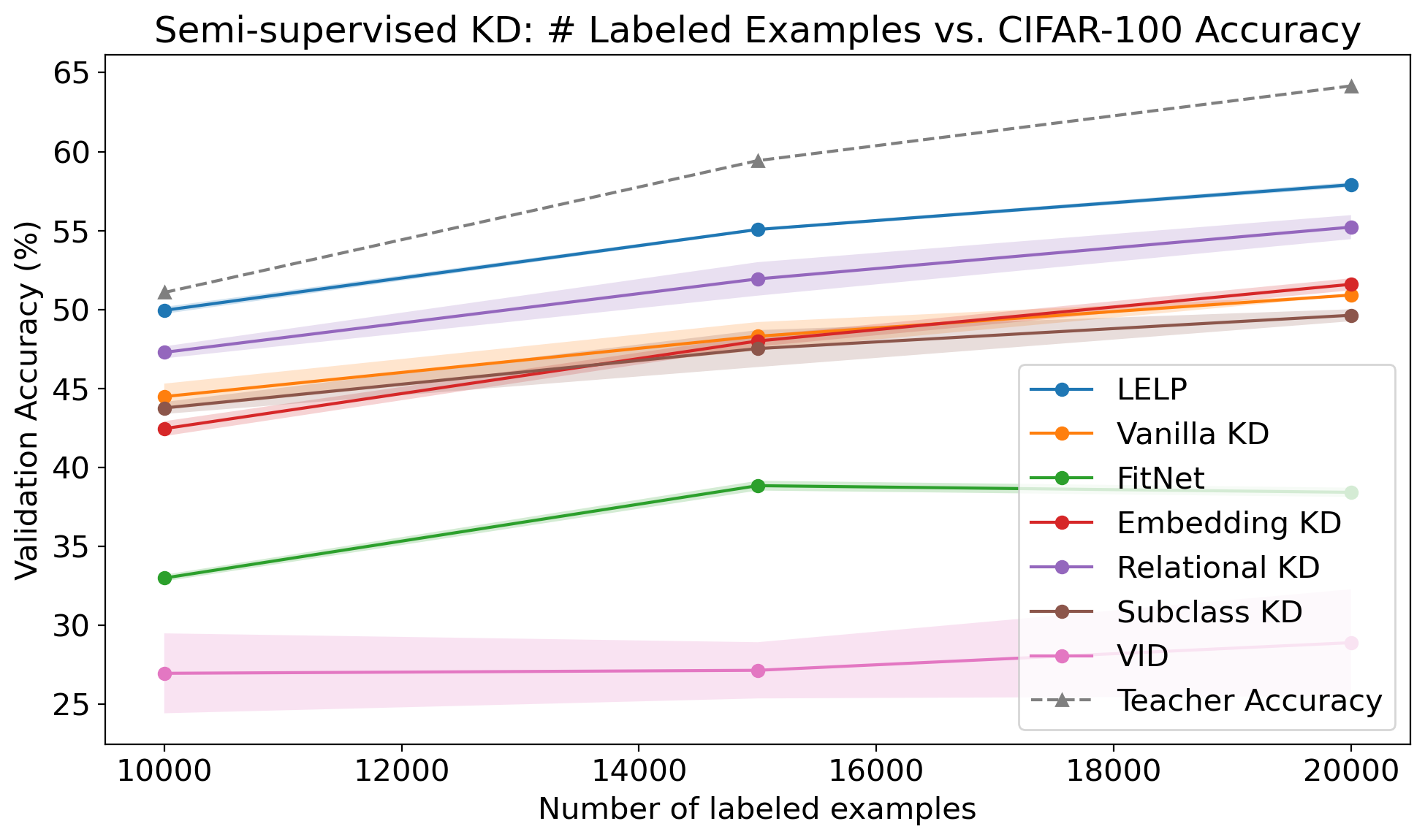}
\caption{Experiments on the binary and standard CIFAR-100 datasets using ResNet92 as the teacher and  ResNet56 and  MobileNet, respectively, as the student. \textbf{Left:} Distillation Dataset Size vs. Accuracy on binary CIFAR-100. LELP achieves the same performance as standard training while using only $25\%$ of the data.  \textbf{Middle:} Student's validation accuracy over the training trajectory. LELP offers both performance gains over Vanilla KD and a faster convergence rate.  \textbf{Right:} Illustration of the performance of LELP  in a semi-supervised setting. The x-axis shows the initial quantity of labeled examples used to train the teacher model, which then generates pseudo-labels for the remaining (unlabeled) portion of the CIFAR-100 dataset.  See Appendix~\ref{app:semi} for more details.}
\label{fig:LELP_properties}
\end{center}
\vskip -0.2in
\end{figure*}

As we demonstrate in Figure~\ref{fig:LELP_properties}, LELP comes with several desirable properties. We consider the binary and standard CIFAR-100 datasets, ResNet92 as the teacher and ResNet56 and MobileNet, respectively, as the student. As we have already discussed in Appendix~\ref{app:multiclass},  we selected CIFAR-10 and CIFAR-100 as our datasets because they are widely recognized benchmarks for 10-class and 100-class classification, respectively, and we intentionally disregarded methods that are tailored to vision-tasks. 

We observe the following. First, the teacher conveys a high amount of information per example, in the sense that LELP is able to achieve the same performance as standard training using only a small fraction of the data. Second,  LELP  both outperforms and converges faster than Vanilla KD. Finally,  in Appendix~\ref{app:semi}  we show  that LELP can offer significant gains in the semi-supervised setting, where the teacher is likely to generate inaccurate pseudo-labels.

\subsection{Semi-supervised KD experiments}
\label{app:semi}

\begin{table*}[h] 
\renewcommand{\arraystretch}{1.15}
\caption{Experiments in the \textbf{semi-supervised KD setting.}}
\label{tab:semi}

\begin{center}
\tiny
\begin{tabular}{|c| c c c |} 
 \hline
 %\rowcolor{cyan}
\cellcolor{cyan}
Teacher Architecture 
& ResNet92 
& ResNet92
& ResNet92
\\ 

 %\hline
 % \rowcolor{cyan}
 \cellcolor{cyan}
Student Architecture& 
MobileNet &  
MobileNet  & 
MobileNet \\ 
% \hline
 \cellcolor{cyan}
Dataset&

CIFAR-100& 
CIFAR-100& 
CIFAR-100\\ 
\cellcolor{cyan}
Number of Labeled Examples&

10000 & 
15000 & 
20000 \\ 
\hline
 \cellcolor[HTML]{FBB982}
 Teacher & 
 $ 51.08$ &  
 $59.43$ & 
 $64.16$  \\
% \hline
 \cellcolor[HTML]{FBB982}
 Subclass Distillation Teacher
 & 
 $ 53.05 $ &  
 $60.22 $ & 
 $63.93 $ \\
\hline
\cellcolor[HTML]{FDBC42}
Standard Training & 
 $ 44.49 \pm 0.82$ &  
 $ 48.31 \pm 0.90 $ &
$ 50.92 \pm 0.04 $  \\
 
 %\hline
\cellcolor[HTML]{FDBC42}
 Vanilla Distillation & 
 $ 44.49 \pm 0.82$ &  
 $ 48.31 \pm 0.90 $ & 
$ 50.92 \pm 0.04 $  \\
 
% \hline
 \cellcolor[HTML]{FDBC42}
 Embedding Distillation &   
 $42.46 \pm 0.47$  & 
 $48.01 \pm 0.35$ & 
 $51.60 \pm 0.38$  \\
 
 %\hline
 \cellcolor[HTML]{FDBC42}
 FitNet &   
 $ 33.0 \pm 0.21$ & 
 $ 38.85 \pm 0.31$ & 
 $ 38.43 \pm 0.29$  \\
 \cellcolor[HTML]{FDBC42}
 VID  & 
 $26.97 \pm 2.53$ & 
 $27.16 \pm 1.78$ & 
 $28.91 \pm 3.38$ \\  
 %\hline
 \cellcolor[HTML]{FDBC42}
  Relational KD  & 
 $\underline{47.29 \pm 0.38}$ & 
  $\underline{51.94 \pm 1.06}$ & 
 $\underline{55.22 \pm 0.76}$ \\  
 %\hline
 \cellcolor[HTML]{FDBC42}
 Subclass Distillation &  
 $43.77 \pm 0.38$ & 
 $47.53 \pm 1.17$ & 
 $49.64 \pm 0.38 $  \\
 %\hline
 \cellcolor[HTML]{FDBC42}
LELP (Ours) &  
 $\mathbf{49.95 \pm 0.22}$ &
 $\mathbf{55.07 \pm 0.09}$ & 
 $\mathbf{57.90 \pm 0.16}$  \\
 \hline
  \cellcolor{lightgray}
 Avg. gain over the best baseline &
\cellcolor{green} $2.66$ & 
\cellcolor{green} $3.13$ & 
\cellcolor{green} $2.68$ \\
  %\hline
\cellcolor{lightgray}
 Avg. gain over non-subclass baseline &
\cellcolor{green} $2.66$ & 
\cellcolor{green} $3.13$ & 
\cellcolor{green} $2.68$  \\
  %\hline
\cellcolor{lightgray}
 Avg. gain over Vanilla KD &
\cellcolor{green} $5.46$ & 
\cellcolor{green} $6.76$ & 
\cellcolor{green} $6.98$  \\
  \hline
  
\end{tabular}

\end{center}

\end{table*}

Semi-supervised KD, also known as KD with unlabeled examples, is a potent training paradigm for generating compact and efficient student models in scenarios where labeled data is scarce but a large pool of unlabeled data exists. This approach employs a high-capacity teacher model to generate (soft) pseudo-labels for the unlabeled dataset, which are subsequently utilized to train the student model. 

Despite its widespread success in practice, the effectiveness of this powerful approach generally depends on the quality of the pseudo-labels generated by the teacher model. Indeed, training the student model on noisy pseudo-labels often leads to significant degradation of its generalization performance, and this is a well-known phenomenon that has been observed and studied in a plethora of papers in the literature, e.g.,~\cite{arazo2020pseudo, liu2021certainty, 
pham2021meta, stanton2021does, xie2020self, RAD, IKBMTV22,kontonis2023slam}. Additionally, enforcing more teacher-student consistency, e.g., by blindly mimicking the teacher's embeddings, can even hinder performance when the teacher's output contains high noise (i.e., one may get worse performance than simply applying Vanilla KD). Finally, in this setting, often times the expense of using the teacher model to generate pseudo-labels can be a significant limitation. This becomes particularly problematic for methods like Subclass Distillation, which necessitate training multiple teacher models and generating pseudo-labels with each one of them, further compounding the cost.

In Table~\ref{tab:semi} (see also  the rightmost plot in Figure~\ref{fig:LELP_properties}) we study such a semi-supervised setting in the case of CIFAR-100, with the teacher model being a ResNet92 and student model being a MobileNet. (We choose this setting because it is the task where the teacher models will be the most noisy, and also the teacher and student models have different architectures.)

We consider the cases where the teacher model has available 10000, 15000, 20000 labeled examples for training, and it is then used to pseudo-label the rest of the CIFAR-100 dataset. (Note that the amount of available labeled examples for training the teacher model directly affects its accuracy.) Finally, the student-model is trained on both the available labeled data (which are also soft-labeled by the teacher model) and the teacher's pseudo-labels on the unlabeled data.

We observe that LELP provides significant gains in this setting. Also notably, for the 10000 and the 15000 case, the only baseline that  outperforms Vanilla KD is Relational KD — showing that, while the teacher's embeddings in a noisy teacher setting may contain valuable information, extracting it effectively is a complex challenge — especially when the teacher and student come from different architecture families.

\section{Detailed Description of LELP}
\label{app:alg_details}

Here we provide further details and pseudo-code for the steps detailed in Section~\ref{sec:our_method}.

\subsection{Step 1: PCA on Teacher Embeddings}

\begin{algorithm}[h]
   \caption{Learning Embedding Linear Projections (LELP) - Step 1 - Computing Subclass Direction}
   \label{alg:embedding_pca}
\begin{algorithmic}
   \STATE {\bfseries Input:} Teacher model feature extractor $h^\text{Teacher}(x)$, teacher final layer weight $W_T\in R^{D\times C}$, dataset with class labels $\{x_i, y_i\}_{i\leq N}$, with class counts $N_c$
   \STATE {\bfseries Output:} $S\times C$ class vectors $\tilde{V} = \{\tilde{v}_{c,s }\}$ and $C$ class means $M = \{\mu_c\}$
    \\\hrulefill
   \STATE Instantiate set of $\tilde{v}_{c,s }$ as empty set $\tilde{V} \leftarrow \{\}$
   \STATE Compute QR Decomposition of $W_T$, $\tilde{W_T} \leftarrow \texttt{QR}(W_T)$ with Gram-Schmidt Algorithm
   \FOR{$c=1$ {\bfseries to} $C$}
        \STATE Get set of all inputs $X^c = \{x_i\}$ belonging to class $c$
        \STATE Compute matrix of teacher features for class $c$, $H^c \in R^{N_c \times D}$, s.t. $H^c_i = h^\text{Teacher}(X^c_i)$
        \STATE Project $H^c$ onto null space of $W_T$: $\tilde{H^c} \leftarrow H^c - H^c \tilde{W}_T$
        \STATE Compute top-$S$ PCA on $\tilde{H}^c$ to obtain $\{v_{c, s}\}$ for $1\leq s \leq S$
        \STATE Produce random orthogonal matrix $Q^c \in R^{S\times S}$
        \STATE Set $\tilde{v}_{c, s} \leftarrow Q^c {v}_{c, s}$ for $1\leq s \leq S$
        \STATE Compute $\{\sigma_{c, s}\}$, as $\sigma_{c, s}^2 = \frac{1}{N_c}||\tilde{H}^c\tilde{v}_{c, s}||_2^2$
        \STATE Normalize $\tilde{v}_{c, s} \leftarrow \frac{\tilde{v}_{c, s} }{max_{s} \{\sigma_{c, s}\}}$ (Normalizing $\tilde{v}_{c, s}$) 
        \STATE Add $\{ \tilde{v}_{c, s} \}$ to $\tilde{V}$: $\tilde{V} \leftarrow \tilde{V} \cup \{ \tilde{v}_{c, s} \}$
   \ENDFOR
   \STATE {\bfseries Return: } $\tilde{V}$
\end{algorithmic}
\end{algorithm}

This corresponds to \cref{sec:methods_pca}, and the main goal of this step is to obtain $S\times C$ vectors $\tilde{v}_{s,c} \in R^D$ to create subclasses from, with $D$ the teacher embedding dimension. The pseudocode is provided in \cref{alg:embedding_pca}. Note that the version provided in \cref{alg:embedding_pca} works with the full teacher embedding matrix for a class $H^c \in R^{N_c \times D}$, but it is straightforward to adapt this to a streaming approach that does not require storing all the embeddings at the same time by keeping running statistics. Note we additionally perform a small normalization step, so that the maximum variance along any of the teacher projection vectors $\{ \tilde{v}_{c, s} \}$ is 1 for any class.

\subsection{Step 2/3: Knowledge Distillation with Subclasses}

\begin{algorithm}[h]
   \caption{Subclass Splitting from teacher embedding, $\texttt{subsplit}(h, \tilde{V}, M, W, \beta, \tau)$}
   \label{alg:pseudo_class_split}
\begin{algorithmic}
   \STATE {\bfseries Input:} Teacher embedding $h$ \\
   \quad\quad\quad $C \times S$ subclass projection vectors $\tilde{V} = \{\tilde{v}_{c,s }\}$ \\ 
   \quad\quad\quad $C$ class means $M = \{\mu_c\}$ \\
    \quad\quad\quad Teacher final layer classification weights $W = [w_1 \dots w_C] \in R^{D\times C}$ and biases $[b_1 \dots b_C]$\\
    \quad\quad\quad Subclass Temperature $\beta$ \\
    \quad\quad\quad Student-Teacher Temperature $\tau$
   \STATE {\bfseries Output:} $C \times S$ Teacher subclass probabilities: $p^{\text{Teacher}}_{c,s }$
    \\\hrulefill\\
    \STATE Compute teacher coarse label logits: $z_c \leftarrow w_c^\intercal h - b_c$
   \STATE Compute teacher $C$ coarse label probabilities with temperature $\tau$: $p_c^\text{Teacher} \leftarrow \frac{e^{z_c/\tau}}{\sum_{j = 1}^C e^{z_c/\tau} }$ ($\tau$ -tempered Softmax)
   \FOR{$c=1$ {\bfseries to} $C$}
        \STATE Compute $S$ subclass logits: $z_{c, s} \leftarrow \tilde{v}_{c, s} ^\intercal (h-\mu_c)$
        \STATE Compute subclass probabilities for class c: 
        $p_{c, s}^\text{Teacher} \leftarrow p_c^\text{Teacher} * \frac{e^{z_{c, s}/\beta}}{\sum_{j = 1}^S e^{z_{c, j}/\beta} }.$ ($\beta$-tempered softmax over subclass logits)
   \ENDFOR
   \STATE {\bfseries Return: } $p_{c, s}^\text{Teacher}$
\end{algorithmic}
\end{algorithm}

\begin{algorithm*}[h]
   \caption{Learning Embedding Linear Projections (LELP) - Knowledge Distillation with Subclasses}
   \label{alg:LELP_train}
\begin{algorithmic}
   \STATE {\bfseries Input:} Teacher model feature extractor $h^\text{Teacher}(x)$\\
   \quad\quad\quad $C \times S$ subclass projection vectors $\tilde{V} = \{\tilde{v}_{c,s }\}$ \\ 
   \quad\quad\quad $C$ class mean vectors $M = \{\mu_c\}$ \\
    \quad\quad\quad Teacher final layer classification weights $W = [w_1 \dots w_C] \in R^{D\times C}$ and biases $[b_1 \dots b_C]$\\
    \quad\quad\quad Subclass Temperature $\beta$ \\
    \quad\quad\quad Student-Teacher Temperature $\tau$ \\
    \quad\quad\quad $C\times S$ class student model $f_{\theta_S}^\text{Student}(x)$ and weights $\theta_S$\\
    \quad\quad\quad Dataset $p(x)$ \\
    \quad\quad\quad Learning rate $\eta$
   \STATE {\bfseries Output:} Trained Student Model $\theta$
   \WHILE {Not converged}
        \STATE Sample $x \sim p(x)$
        \STATE Compute teacher embedding $h^\text{Teacher} \leftarrow h^\text{Teacher}(x)$
        \STATE Compute teacher subclass probabilities $p^\text{Teacher}_{s,c} \leftarrow \texttt{subsplit}(h, \tilde{V}, M, W, \beta, \tau)$
        \STATE Compute $C\times S$ student logits: $z^\text{Student}_{c,s} \leftarrow f^\text{Student}_{\theta_S}(x)$
        \STATE Compute $C\times S$ tempered student probabilities: $p_{c, s}^\text{Student} \leftarrow \frac{e^{z^\text{Student}_{c, s}/\tau}}{\sum_{i = 1}^C\sum_{j = 1}^S e^{z^\text{Student}_{i, j}/\tau}}$
        \STATE Compute LELP loss $\mathcal{L}_{LELP} = \tau^2 \sum_{c = 1}^C\sum_{s = 1}^S p^\text{Teacher}_{c,s} \log\frac{p^\text{Teacher}_{c,s}}{p^\text{Student}_{c,s}}$ (Standard KL Divergence)
        \STATE Update student parameters: $\theta_S \leftarrow \theta_S - \eta \frac{\partial \mathcal{L}_{LELP}}{\partial \theta_S}$
   \ENDWHILE
   \STATE {\bfseries Return: } $\theta_S$
\end{algorithmic}
\end{algorithm*}

This corresponds to Section~\ref{sec:methods_pseudo_class_splitting} and Section~\ref{sec:methods_kd_with_subclasses}. In Algorithm~\ref{alg:pseudo_class_split} we describe how to generate the pseudo-subclasses from a given teacher embedding, $h$. This subprocess is used in Algorithm~\ref{alg:LELP_train} for knowledge distillation. In Algorithm~\ref{alg:LELP_train} we describe the pure-knowledge distillation setting (i.e. with $\alpha = 0$ in \cref{student_loss}), meaning we are not using the hard labels at all, as we did in the main text, but it is straightforward to combine with with the standard cross-entropy loss with hard labels using \cref{student_loss}.

\section{Implementation Details}
\label{sec:implementation_details}

We implemented all algorithms in Python and used the TensorFlow deep learning library~\cite{abadi2016tensorflow}. We ran our experiments on 64 Cloud TPU v4s each with two cores.

For a fair comparison, we use the teacher's last-layer embeddings throughout the distillation process across all relevant baselines (Embedding Distillation, FitNet, VID, Relational KD, CRD). Furthermore, to handle mismatches in embedding dimensions between teacher and student, we introduce a trainable fully connected layer as a learnable projection for all these methods. (For Vision datasets, we have experimented with convolutional layers as suggested in~\cite{romero2014fitnets}, and while these do reduce the size of the student model, they do not significantly impact its performance compared to the fully connected ones.) 

For FitNet and CRD, in any given experiment, we pre-train the learnable projection once. This same pre-trained projection is then utilized consistently across all three trials corresponding to the experiment. The optimizer used for the pre-training of the learnable projection is Adam with initial learning rate $10^{-3}$ and $10^{-6}$ for Vision ($200$ epochs training) and Natural Language ($40$ epochs training) datasets, respectively. 

We implement VID by using the loss function as described in (5) of~\cite{Ahn2019CVPR} where the squared difference in the second term is taken over the teacher's and student's embeddings (using a learnable projection if there is a mismatch in their dimensions).

We implement Relational KD using the loss function as described in (9) and (10) of ~\cite{park2019relational}.

For CRD we implement the  objective described in (17) of~\cite{tian2019contrastive} and perform grid search over $\{0.1, 1.0, 5.0, 10.0, 100.0\}$ for its coefficient. We consider the standard negative sampling policy where a pair of examples $(x,y)$ is considered ``negative" if $x \ne y$, i.e., given a batch the number of positive pairs is equal to the batch size. (Note that in Vision classification tasks the standard approach would be to generate variations of every image (e.g, via rotations), and then also consider as positive any pair of examples (original or modified) that come from the same source. However, data augmentation of Natural Language examples is not straightforward and this impairs the performance of the method — this is the main point we are trying to convey here.)

For DKD we implement equations (1), (2) and (7) in~\cite{zhao2022decoupled}. We set the 'target'  hyperparameter $\alpha$ equal to $1$ and perform grid search for the ``non-target" hyperparameter $\beta$ over $\{1, 2, 4,  6, 8, 10\}$ (following examples in~\cite{zhao2022decoupled}).

The hyperparameters chosen for each method and dataset can be found in Appendix~\ref{hyperparameters_chosen}.

% For the CRD and DKD implementation details the reader is referred to Appendix~\ref{app:DKD_CRD}

\subsection{Vision Datasets}

In every experiment, both the teacher and student models are trained for $200$ epochs.

For training ResNet-92 we use the SGD optimizer with initial learning rate $10^{-3}$, non-Nesterov momentum value equal to $0.9$, cosine annealing learning rate schedule (minimum learning rate value is set to $10^{-6}$), and batch size $256$.  For training ResNet-56 we use the SGD optimizer with initial learning rate $5 \cdot 10^{-2}$, Nesterov momentum value equal to $0.9$, cosine annealing learning rate schedule (minimum learning rate value is set to $10^{-6}$) and batch size $256$. (This is a training schedule similar to the one described in~\cite{stanton2021does}).

For training MobileNet (both the larger teacher-model and the student-model) we use the Adam optimizer with initial learning rate $\mathrm{lr} = 0.001$ and batch size $128$.  We then proceed according to the following learning rate schedule for $200$ epochs  (see, e.g.,~\cite{he2016deep}):
\begin{align*}
\mathrm{lr} \leftarrow
\begin{cases}
\mathrm{lr} \cdot 0.5 \cdot 10^{-3}, & \text{if  $\#\mathrm{epochs} > 180$ } \\
\mathrm{lr}  \cdot 10^{-3}, & \text{if  $\#\mathrm{epochs} > 160$ } \\
\mathrm{lr}  \cdot 10^{-2}, & \text{if  $\#\mathrm{epochs} > 120$ } \\
\mathrm{lr}  \cdot 10^{-1}, & \text{if  $\#\mathrm{epochs} > 80$ }
\end{cases}
\end{align*}

Finally, in all cases we use data-augmentation. In particular, we use random horizontal flipping and random width and height translations with width and height factor, respectively, equal to $0.1$.

\subsection{Natural Language Datasets}
\label{app:NLP_details}

For the GLUE/COLA dataset, the teacher model undergoes training for $2$ epochs. In the case of GLUE/SST-2 and the Large Movie Review Dataset, teacher model training extends to $3$ epochs.  We employ the Adam optimizer (batch size $64$, initial learning rate $10^{-5}$) for these training processes. To minimize variance across experiments, we consistently train the student model for $40$ epochs using a smaller learning rate. The training uses the Adam optimizer with a batch size of $64$ and an initial learning rate of $10^{-6}$.

For both the Amazon Reviews datasets, the teacher and student models (both the ALBERT-base and the MLPs over frozen-T5 embeddings) are trained for $2$ epochs using the Adam optimizer with a batch size of $64$ and an initial learning rate of $10^{-6}$.

For the Sentiment140 Bin dataset the teacher and the student models are trained for $1$ epoch using the Adam optimizer with batch size of $64$ and an initial learning rate of $10^{-6}$.

\subsection{Hyperparameter optimization}
\label{hyperparameters_chosen}

In this section, we present the hyperparameter optimization (grid search) procedure we followed for each method, dataset and experiment of Section~\ref{sec:experimental_evaluation} and Appendix~\ref{unsupervised_clustering_comparison}. For Vision datasets, \textbf{bold} numbers correspond to the hyperparameters chosen for the case of  ResNet56 student,  the number in \emph{italicized} numbers correspond to the hyperparameters chosen for the case of MobileNet student with ResNet92 as a teacher and, finally, the \underline{underlined} numbers correspond to the case of MobileNet student with (a larger) MobileNet as a teacher. For Natural Language datasets,  \textbf{bold} numbers correspond to the case of ALBERT-base student, and \emph{italicized} numbers correspond to the case of a MLP over frozen sentence-T5 (11B) embeddings as a student.

\begin{table*}[h!] 
\renewcommand{\arraystretch}{1.15}

\label{tab:hyperparameters_vanilla_embedding}

\begin{minipage}{0.37\textwidth}

\centering
\caption{Hyperparameters for Vanilla KD}
\tiny

\resizebox{1.0\textwidth}{!}{%
\begin{tabular}{|c| c |} 
 \hline
 %\rowcolor{cyan}
Dataset
& Temperature \\ 

 \hline
CIFAR-10-bin
& $\{\textbf{1.0}, 2.0, 3.0, \underline{4.0}, 5.0, \emph{10.0}\}$ \\ 

% \hline
 % \rowcolor{cyan}

CIFAR-100-bin& 
 $\{\textbf{\underline{1.0}}, 2.0, 3.0, 4.0, 5.0, \emph{10.0}\}$  \\ 
%\hline

CIFAR-10& 
 $\{1.0, \textbf{2.0}, 3.0, 4.0, \underline{\emph{5.0}}, 10.0\}\}$  \\ 
%\hline

CIFAR-100& 
 $\{1.0, 2.0, 3.0, 4.0, 5.0, \textbf{\emph{10.0}}\}$  \\ 
%\hline

LMRD& 
 $\{\textbf{1.0}, 2.0, 3.0, 4.0, 5.0, 10.0\}$  \\ 
%\hline

GLUE/cola& 
 $\{1.0, 2.0, 3.0, \textbf{4.0}, 5.0, 10.0\}$  \\ 
%\hline

GLUE/sst2& 
 $\{1.0, 2.0, \textbf{3.0}, 4.0, 5.0, 10.0\}$  \\

Amazon Reviews Bin& 
 $\{\emph{1.0}, 2.0, 3.0, \textbf{4.0}, 5.0, 10.0\}$  \\ 

 Amazon Reviews & 
 $\{\textbf{1.0}, \emph{2.0}, 3.0, 4.0, 5.0, 10.0\}$  \\ 

Sentiment140 Bin & 
 $\{\textbf{1.0}, 2.0, 3.0, 4.0, 5.0, 10.0\}$  \\

\hline
\end{tabular}
}
\end{minipage}
\begin{minipage}{0.61\textwidth}

\centering
\caption{Hyperparameters for Embedding KD}
\tiny

\resizebox{1.0\textwidth}{!}{%
\begin{tabular}{|c| c c|} 

\hline
 %\rowcolor{cyan}
Dataset
& Temperature
& Embeddings-loss coefficient
\\ 

 \hline
CIFAR-10-bin
& $\{\textbf{1.0}, 2.0, 3.0, \underline{4.0}, 5.0, \emph{10.0}\}$  
& $\{\underline{0.1}, \emph{1.0}, 5.0, 10.0, \textbf{100.0}, 1000.0 \}$ \\ 

% \hline
 % \rowcolor{cyan}

CIFAR-100-bin
&  $\{\textbf{1.0},\underline{2.0}, 3.0, 4.0, 5.0, \emph{10.0}\}$  
 & $\{ \underline{\emph{0.1}}, 1.0, 5.0, 10.0, \textbf{100.0}, 1000.0 \}$ \\ 
%\hline

CIFAR-10 
&  $\{\underline{1.0}, \textbf{2.0}, 3.0, 4.0, 5.0, \emph{10.0}\}$  
 & $\{\emph{0.1}, 1.0, \underline{5.0}, 10.0, \textbf{100.0}, 1000.0 \}$ \\ 
%\hline

CIFAR-100
&  $\{\textbf{1.0}, 2.0, 3.0, 4.0, 5.0, \emph{10.0}\}$  
 & $\{\emph{0.1}, \textbf{1.0}, 5.0, 10.0, 100.0, 1000.0 \}$ \\ 
%\hline

LMRD
&  $\{1.0, 2.0, 3.0, \textbf{4.0}, 5.0, 10.0\}$  
 & $\{\textbf{0.1}, 1.0, 5.0, 10.0, 100.0, 1000.0 \}$ \\ 
%\hline

GLUE/cola 
&  $\{1.0, \textbf{2.0}, 3.0, 4.0, 5.0, 10.0\}$  
 & $\{0.1, \textbf{1.0}, 5.0, 10.0, 100.0, 1000.0 \}$ \\ 
%\hline

GLUE/sst 
&  $\{1.0, 2.0, \textbf{3.0}, 4.0, 5.0, 10.0\}$  
 & $\{\textbf{0.1}, 1.0, 5.0, 10.0, 100.0, 1000.0 \}$ \\

Amazon Reviews Bin 
&  $\{1.0, 2.0, 3.0, \textbf{\emph{4.0}}, 5.0, 10.0\}$  
 & $\{\textbf{\emph{0.1}}, 1.0, 5.0, 10.0, 100.0, 1000.0 \}$ \\

Amazon Reviews
&  $\{\emph{\textbf{1.0}}, 2.0, 3.0, 4.0, 5.0, 10.0\}$  
 & $\{\emph{0.1}, 1.0, 5.0, \textbf{10.0}, 100.0, 1000.0 \}$ \\

Sentiment140 Bin
&  $\{1.0, 2.0, \textbf{3.0}, 4.0, 5.0, 10.0\}$  
 & $\{\textbf{0.1}, 1.0, 5.0, 10.0, 100.0, 1000.0 \}$ \\

\hline

\end{tabular}
}
\end{minipage}

\end{table*}

\begin{table*}[h!] 
\renewcommand{\arraystretch}{1.15}

\label{tab:hyperparameters_fitnet}

\centering
\caption{Hyperparameters for FitNet}
\tiny
\begin{tabular}{|c| c c|} 
 \hline
 %\rowcolor{cyan}
Dataset
& Temperature
& Embeddings-loss coefficient
\\ 

 \hline
CIFAR-10-bin
& $\{ \textbf{1.0}, 2.0, 3.0, \underline{4.0}, \emph{5.0}, 10.0\}$  
& $\{0.1, \underline{1.0}, 5.0, 10.0, \textbf{100.0}, \emph{1000.0} \}$ \\ 

% \hline
 % \rowcolor{cyan}

CIFAR-100-bin
&  $\{\underline{\textbf{1.0}}, 2.0, 3.0, 4.0, 5.0, \emph{10.0}\}$  
 & $\{ 0.1, \underline{\emph{1.0}}, 5.0, 10.0, \textbf{100.0}, 1000.0 \}$ \\ 
%\hline

CIFAR-10 
&  $\{\textbf{1.0}, 2.0, 3.0, \emph{4.0}, 5.0, \underline{10.0}\}$  
 & $\{0.1, 1.0, \emph{5.0}, \underline{10.0}, \textbf{100.0}, 1000.0 \}$ \\ 
%\hline

CIFAR-100
&  $\{\textbf{1.0}, 2.0, 3.0, 4.0, \emph{5.0}, 10.0\}$  
 & $\{\emph{0.1}, \textbf{1.0}, 5.0, 10.0, 100.0, 1000.0 \}$ \\ 
%\hline

LMRD
&  $\{ \textbf{1.0}, 2.0, 3.0, 4.0, 5.0, 10.0\}$  
 & $\{0.1, \textbf{1.0}, 5.0, 10.0, 100.0, 1000.0 \}$ \\ 
%\hline

GLEU/cola 
&  $\{1.0, 2.0, 3.0, 4.0, \textbf{5.0}, 10.0\}$  
 & $\{0.1, 1.0, 5.0, \textbf{10.0}, 100.0, 1000.0 \}$ \\ 
%\hline

GLEU/sst 
&  $\{1.0, 2.0, \textbf{3.0}, 4.0, 5.0, 10.0\}$  
 & $\{0.1, 1.0, \textbf{5.0}, 10.0, 100.0, 1000.0 \}$ \\ 

Amazon Reviews Bin
&  $\{1.0, \textbf{2.0}, 3.0, 4.0, 5.0, \emph{10.0}\}$  
 & $\{0.1, \textbf{1.0}, 5.0, \emph{10.0}, 100.0, 1000.0 \}$ \\ 
%\hline

Amazon Reviews
&  $\{1.0, 2.0, \textbf{3.0}, 4.0, \textbf{5.0}, 10.0\}$  
 & $\{\emph{0.1}, 1.0, 5.0, \textbf{10.0}, 100.0, 1000.0 \}$ \\

Sentiment140 Bin
&  $\{1.0, 2.0, \textbf{3.0}, 4.0, 5.0, 10.0\}$  
 & $\{\textbf{0.1}, 1.0, 5.0, 10.0, 100.0, 1000.0 \}$ \\

\hline

\end{tabular}
\end{table*}
\begin{table*}[h!] 
\renewcommand{\arraystretch}{1.15}

\label{tab:hyperparameters_vid}

\centering
\caption{Hyperparameters for VID}
\tiny

\begin{tabular}{|c| c c|} 
 \hline
 %\rowcolor{cyan}
Dataset
& Temperature
& Embeddings-loss coefficient
\\ 

 \hline
CIFAR-10-bin
& $\{\underline{\textbf{1.0}}, 2.0, 3.0, 4.0, 5.0, \emph{10.0}\}$  
& $\{ \emph{0.1}, \textbf{0.2},  0.5, 1.0, \underline{5.0}, 10.0,  100.0 \}$ \\ 

% \hline
 % \rowcolor{cyan}

CIFAR-100-bin
& $\{1.0, \textbf{2.0}, 3.0, 4.0, \emph{5.0}, \underline{10.0}\}$  
& $\{ \textbf{0.1}, 0.2,  \emph{0.5}, 1.0, 5.0, 10.0,  \underline{100.0} \}$ \\ 

%\hline

CIFAR-10 
& $\{1.0, 2.0, 3.0, \textbf{4.0}, \underline{5.0}, \emph{10.0}\}$  
& $\{ \underline{\textbf{\emph{0.1}}}, 0.2,  0.5, 1.0, 5.0, 10.0,  100.0 \}$ \\ 
%\hline

CIFAR-100
& $\{1.0, 2.0, \emph{3.0}, 4.0, 5.0, \textbf{10.0}\}$  
& $\{ \textbf{0.1}, 0.2,  0.5, \emph{1.0}, 5.0, 10.0,  100.0 \}$ \\ 
%\hline

LMRD
& $\{ 1.0, 2.0, 3.0, 4.0, 5.0, \textbf{10.0}\}$  
& $\{ \textbf{0.1}, 0.2,  0.5, 1.0, 5.0, 10.0,  100.0 \}$ \\ 
%\hline

GLEU/cola 
& $\{1.0, \textbf{3.0}, 3.0, 4.0, 5.0, 10.0\}$  
& $\{ \textbf{0.1}, 0.2,  0.5, 1.0, 5.0, 10.0,  100.0 \}$ \\ 
%\hline

GLEU/sst 
& $\{1.0, 2.0, 3.0, 4.0, 5.0, \textbf{10.0}\}$  
& $\{ \textbf{0.1}, 0.2,  0.5, 1.0, 5.0, 10.0,  100.0 \}$ \\

Amazon Reviews Bin
& $\{1.0, 2.0, 3.0, \emph{4.0}, 5.0, \textbf{10.0}\}$ 
& $\{\textbf{0.1}, 0.2,  0.5, 1.0, 5.0, 10.0,  \emph{100.0} \}$ \\ 
%\hline

Amazon Reviews
& $\{1.0, 2.0, 3.0, \emph{4.0}, 5.0, \textbf{10.0}\}$  
& $\{ \textbf{0.1}, 0.2,  0.5, 1.0, \emph{5.0}, 10.0,  100.0 \}$ \\ 

Sentiment140 Bin
&  $\{\textbf{1.0}, 2.0, 3.0, 4.0, 5.0, 10.0\}$  
 & $\{\textbf{0.1}, 1.0, 5.0, 10.0, 100.0, 1000.0 \}$ \\

\hline

\end{tabular}

\end{table*}
\begin{table*}[h!] 
\renewcommand{\arraystretch}{1.15}

\label{tab:hyperparameters_relational}

\centering
\caption{Hyperparameters for Relational KD}
\tiny

\begin{tabular}{|c| c c|} 
 \hline
 %\rowcolor{cyan}
Dataset
& Temperature
& Embeddings-loss coefficient
\\ 

 \hline
CIFAR-10-bin
& $\{1.0, \textbf{\underline{2.0}}, 3.0, 4.0, 5.0, \emph{10.0}\}$  
& $\{\emph{0.1}, 0.2,  0.5, \textbf{1.0}, 5.0, 10.0,  \underline{100.0} \}$ \\ 

% \hline
 % \rowcolor{cyan}

CIFAR-100-bin
& $\{\textbf{1.0}, \underline{2.0}, 3.0, 4.0, 5.0, \emph{10.0}\}$  
& $\{ \emph{0.1}, 0.2, 0.5, 1.0, 5.0, \underline{\textbf{10.0}},  100.0 \}$ \\ 

%\hline

CIFAR-10 
& $\{\textbf{1.0}, 2.0, 3.0, 4.0, 5.0, \underline{\emph{10.0}}\}$  
& $\{ 0.1, 0.2,  0.5, 1.0, 5.0, 10.0,  \underline{\textbf{\emph{100.0}}} \}$ \\ 
%\hline

CIFAR-100
& $\{1.0, 2.0, 3.0, 4.0, 5.0, \textbf{\emph{10.0}}\}$  
& $\{ \textbf{0.1}, 0.2,  0.5, 1.0, 5.0, 10.0,  \emph{100.0} \}$ \\ 
%\hline

LMRD
& $\{ \textbf{1.0}, 2.0, 3.0, 4.0, 5.0, 10.0\}$  
& $\{ 0.1, 0.2,  0.5, \textbf{1.0}, 5.0, 10.0,  100.0 \}$ \\ 
%\hline

GLEU/cola 
& $\{1.0, \textbf{2.0},  3.0, 4.0, 5.0, 10.0\}$  
& $\{ \textbf{0.1}, 0.2,  0.5, 1.0, 5.0, 10.0,  100.0 \}$ \\ 
%\hline

GLEU/sst 
& $\{\textbf{1.0}, 2.0, 3.0, 4.0, 5.0, 10.0\}$  
& $\{ 0.1, 0.2,  0.5, \textbf{1.0}, 5.0, 10.0,  100.0 \}$ \\

Amazon Reviews Bin
& $\{\textbf{1.0}, 2.0, 3.0, \emph{4.0}, 5.0, 10.0\}$ 
& $\{0.1, 0.2,  0.5, 1.0, \emph{5.0}, \textbf{10.0}, 100.0 \}$ \\ 
%\hline

Amazon Reviews
& $\{\textbf{1.0}, 2.0, 3.0, \emph{4.0}, 5.0, 10.0\}$  
& $\{ 0.1, 0.2,  0.5, \textbf{1.0}, 5.0, \emph{10.0},  100.0 \}$ \\

Sentiment140 Bin
&  $\{\textbf{1.0}, 2.0, 3.0, 4.0, 5.0, 10.0\}$  
 & $\{\textbf{0.1}, 1.0, 5.0, 10.0, 100.0, 1000.0 \}$ \\

\hline

\end{tabular}

\end{table*}
\begin{table*}[h!] 
\renewcommand{\arraystretch}{1.15}

\label{tab:hyperparameters_subclass}

\centering
\caption{Hyperparameters for Subclass Distillation}
\tiny

\resizebox{1.0\textwidth}{!}{%
\begin{tabular}{|c| c c c c|} 
 \hline
 %\rowcolor{cyan}
Dataset
& Num. Subclasses
& Auxiliary loss Temp.
& Auxiliary loss weight
&Distill. Temp.  \\ 

 \hline
CIFAR-10-bin
& $\{2, 3, 4, 5 , \underline{\textbf{\emph{10}}}\}$ 
& $ \{1.0, \textbf{5.0}, \underline{\emph{10.0}} \} $
& $ \{\underline{0.0}, \textbf{\emph{0.1}}, 1.0, 5.0, 10.0, 100.0 \}$
& $ \{1.0, \textbf{2.0}, \emph{3.0}, 4.0, 5.0, \underline{10.0} \}$
\\ 
%\hline

CIFAR-100-bin
& $\{\underline{5}, 10, 15, 20, 25, 30, \emph{35}, 40, 45, 50, \textbf{75}, 100\}$ 
& $ \{\underline{\textbf{1.0}}, 5.0, \emph{10.0} \} $
& $ \{\underline{\textbf{\emph{0.0}}}, 0.1, 1.0,  5.0, 10.0, 100.0 \}$
& $ \{\underline{\textbf{1.0}}, \emph{2.0}, 3.0, 4.0, 5.0, 10.0 \}$
\\ 
%\hline

CIFAR-10
& $\{ \emph{2}, \textbf{4}, 8, \underline{10} \}$ 
& $ \{\underline{1.0}, \textbf{\emph{5.0}}, 10.0 \} $
& $ \{\underline{\emph{0.0}}, \textbf{0.1}, 1.0,  5.0, 10.0, 100.0 \}$
& $ \{1.0, \underline{\textbf{2.0}}, \emph{3.0}, 4.0, 5.0, 10.0 \}$
\\ 
%\hline

CIFAR-100
& $\{ \textbf{\emph{2}}, 4, 8, 10\}$ 
& $ \{\emph{1.0}, \textbf{5.0}, 10.0 \} $
& $ \{0.0, 0.1, \textbf{1.0},  5.0,  \emph{10.0}, 100.0 \}$
& $ \{1.0, 2.0, \textbf{\emph{3.0}}, 4.0, 5.0, 10.0 \}$\\
%\hline
LMRD
& $\{ 2, 3, 4, 5 , \textbf{10}, 25, 50\}$ 
& $ \{1.0, \textbf{5.0}, 10.0 \} $
& $ \{0.0, \textbf{0.1}, 1.0, 5.0, 10.0, 100.0 \}$
& $ \{\textbf{1.0}, 2.0, 3.0, 4.0, 5.0, 10.0 \}$
\\ 
%\hline
GLEU/cola
& $\{2, 3, \textbf{4}, 5 , 10\}$ 
& $ \{1.0, 5.0, \textbf{10.0} \} $
& $ \{\textbf{0.0}, 0.1, 5.0, 10.0, 100.0 \}$
& $ \{1.0, 2.0, 3.0, 4.0, \textbf{5.0}, 10.0 \}$
\\ 
%\hline
GLEU/sst2
& $\{2, 3, 4, 5, \textbf{10}\}$ 
& $ \{1.0, \textbf{5.0}, 10.0 \} $
& $ \{0.0, \textbf{0.1}, 5.0, 10.0, 100.0 \}$
& $ \{\textbf{1.0}, 2.0, 3.0, 4.0, 5.0, 10.0 \}$
\\ 
Amazon Reviews Bin
& $\{2, \emph{\textbf{4}}, 5, 10, 20, 40, 50, 100, 200, 500\}$ 
&   $ \{\emph{\textbf{1.0}}, 5.0, 10.0 \} $
& $ \{\emph{\textbf{0.0}}, 0.1, 5.0, 10.0, 100.0 \}$
& $ \{\emph{1.0}, \textbf{2.0}, 3.0, 4.0, 5.0, 10.0 \}$
\\ 
%\hline
Amazon Reviews 
& $\{\emph{\textbf{2}}, 4, 8, 16, 20, 40, 80, 200\}$ 
& $ \{\textbf{1.0}, 5.0, 10.0 \} $
& $ \{\emph{0.0}, 0.1, \emph{\textbf{5.0}}, 10.0, 100.0 \}$
& $ \{\emph{\textbf{1.0}}, 2.0, 3.0, 4.0, 5.0, 10.0 \}$
\\

Sentiment140 Bin
& $\{\textbf{2}, 4, 5, 10, 20, 40, 50, 100, 200, 500\}$ 
&   $ \{ \textbf{1.0}, 5.0, 10.0 \} $
& $ \{ \textbf{0.0}, 0.1, 5.0, 10.0, 100.0 \}$
& $ \{\textbf{1.0}, 2.0, 3.0, 4.0, 5.0, 10.0 \}$
\\ 
%\hline

\hline
\end{tabular}
}

\end{table*}
\begin{table*}[h!] 
\renewcommand{\arraystretch}{1.15}

\label{tab:hyperparameters_lelp}

\centering
\caption{Hyperparameters for LELP (ours)}
\tiny

\resizebox{1.0\textwidth}{!}{%
\begin{tabular}{|c| c  c c|} 
 \hline
 %\rowcolor{cyan}
Dataset
& Num. Subclasses
& Subclass Temp.
&Distill. Temp.  \\ 

 \hline
CIFAR-10-bin
& $\{5, 10 , \underline{\textbf{\emph{20}}}\}$ 
& $ \{\underline{1/32}, \emph{1/16}, 1/8, \textbf{1/4}, 1/2, 1 \} $
& $ \{ \textbf{1.0}, 2.0, \underline{\emph{4.0}}, 8.0, 10.0 \}$
\\ 
%\hline

CIFAR-100-bin
& $\{5, \underline{10} , 15, \textbf{20}, 25, 30, 35, 40, 45, \emph{50}, 75, 100\}$ 
& $ \{\underline{1/32}, 1/16, 1/8, \emph{1/4}, \textbf{1/2}, 1 \} $
& $ \{1.0, \textbf{2.0}, \underline{\emph{4.0}}, 8.0, 10.0 \}$
\\ 
%\hline

CIFAR-10
& $\{2 , 4, \textbf{8}, \underline{\emph{10}}\}$ 
& $ \{1/32, 1/16, 1/8, 1/4, \textbf{\emph{1/2}}, \underline{1} \} $
& $ \{1.0, 2.0, \underline{\textbf{4.0}}, 8.0, \emph{10.0} \}$
\\ 
%\hline

CIFAR-100
& $\{\textbf{2}, 4, \emph{8}, 10\}$ 
& $ \{1/32, 1/16, 1/8, 1/4, 1/2, \textbf{\emph{1}} \} $
& $ \{1.0, 2.0, 4.0, \textbf{8.0}, \emph{10.0} \}$
\\ 
%\hline
LMRD
& $\{5, \textbf{10} , 15,  20\}$ 
& $ \{1/32, 1/16, 1/8, 1/4, 1/2, \textbf{1} \} $
& $ \{ \textbf{1.0}, 2.0, 3.0,  4.0, 5.0,  10.0 \}$
\\ 
%\hline
GLEU/cola
& $\{5, 10, \textbf{15}, 20\}$ 
& $ \{1/32, 1/16, 1/8, \textbf{1/4}, 1/2, 1 \} $
& $ \{1.0, 2.0, \textbf{3.0}, 4.0, 5.0,  10.0 \}$
\\ 
%\hline
GLEU/sst2
& $\{\textbf{5}, 10, 15, 20\}$ 
& $ \{1/32, 1/16, 1/8, 1/4, \textbf{1/2}, 1 \} $
& $ \{1.0, \textbf{2.0}, 3.0, 4.0, 5.0,  10.0 \}$
\\ 
Amazon Reviews Bin
& $\{2, \emph{4}, 5, 10, 20, 40, 50, \textbf{100}, 200, 500\}$ 
& $ \{1/32, \textbf{1/16}, \emph{1/8}, 1/4, 1/2, 1 \} $
& $ \{ \emph{1.0}, 2.0, 3.0, 4.0, \textbf{5.0},  10.0 \}$
\\ 
Amazon Reviews
& $\{2, 4, 8, 16, 20, 40, \emph{80}, \textbf{200}\}$ 
& $ \{1/32, 1/16, \textbf{1/8}, 1/4, 1/2, \emph{1} \} $
& $ \{ \emph{1.0}, \textbf{2.0}, 3.0, 4.0, 5.0,  10.0 \}$ \\

Sentiment140 Bin
& $\{2, 4, 5, 10, 20, 40, 50, 100, 200, \textbf{500}\}$ 
& $ \{1/32, 1/16, 1/8, 1/4, 1/2, \textbf{1} \} $
& $ \{ 1.0, \textbf{2.0}, 3.0, 4.0, 5.0,  10.0 \}$ \\

\hline
\end{tabular}
}

\end{table*}
% \input{tables/hyperparams_agglo}
% \vspace{-5cm}
% \input{tables/hyperparams_kmeans}
% \vspace{-5cm}
% \input{tables/hyperparams_tsne_kmeans}
 \begin{table*}[h!]
\renewcommand{\arraystretch}{1.15}

\label{tab:hyperparameters_clusterings}

\centering
% \vspace{-5cm}
\caption{Hyperparameters for Agglomerative Clustering}
\tiny

\begin{tabular}{|c| c |} 
 \hline
 %\rowcolor{cyan}
Dataset
& Number of Clusters \\ 

 \hline
CIFAR-10-bin
& $\{ \underline{\textbf{4}}, 6, 8, \emph{10}, 20, 50, 100, 1000\}$ \\ 

% \hline
 % \rowcolor{cyan}

CIFAR-100-bin
& $\{ \underline{\textbf{10}}, 20, 30, 40, 50, 60, \emph{70}, 80, 90, 100, 150, 200\}$ \\ 
\hline

\end{tabular}

\vspace{1cm}

\centering
\caption{Hyperparameters for K-Means Clustering}
\tiny
\begin{tabular}{|c| c |} 
 \hline
 %\rowcolor{cyan}
Dataset
& Number of Clusters \\ 

 \hline
CIFAR-10-bin
& $\{ 4, \underline{\emph{6}}, 8, \textbf{10}, 20, 50, 100, 1000\}$ \\ 

% \hline
 % \rowcolor{cyan}

CIFAR-100-bin
& $\{ \underline{\textbf{10}}, 20, 30, 40, \emph{50}, 60, 70, 80, 90, 100, 150, 200\}$ \\ 
\hline

\end{tabular}

\vspace{1cm}

\caption{Hyperparameters for t-SNE \& K-Means Clustering}
\tiny
\begin{tabular}{|c| c |} 
 \hline
 %\rowcolor{cyan}
Dataset
& Number of Clusters \\ 

 \hline
CIFAR-10-bin
& $\{ 2, \textbf{3}, \underline{\emph{4}}, 5, 10, 25, 50, 500\}$ \\ 

% \hline
 % \rowcolor{cyan}

CIFAR-100-bin
& $\{ \textbf{\emph{5}}, 10, 15, 20, 25, 30, 35, 40, 45, \underline{50}, 75, 100\}$ \\ 
\hline

\end{tabular}

\end{table*}

% \centering
% \begin{table*}[h!]

% \caption{Hyperparameters for t-SNE \& K-Means Clustering}
% \tiny
% \begin{tabular}{|c| c |} 
%  \hline
%  %\rowcolor{cyan}
% Dataset
% & Number of Clusters \\ 

%  \hline
% CIFAR-10-bin
% & $\{ 2, \textbf{3}, \emph{4}, 5, 10, 25, 50, 500\}$ \\ 

% % \hline
%  % \rowcolor{cyan}

% CIFAR-100-bin
% & $\{ \textbf{\emph{5}}, 10, 15, 20, 25, 30, 35, 40, 45, 50, 75, 100\}$ \\ 
% \hline

% \end{tabular}
% \end{table*}

\end{document}